\documentclass[lettersize,journal]{IEEEtran}
\usepackage{amsmath,amsfonts}
\usepackage{algorithmic}
\usepackage{algorithm}
\usepackage{array}
\usepackage[caption=false,font=normalsize,labelfont=sf,textfont=sf]{subfig}
\usepackage{textcomp}
\usepackage{stfloats}
\usepackage{url}
\usepackage{verbatim}
\usepackage{graphicx}
\usepackage{cite}
\usepackage{hyperref}
\usepackage{booktabs}
\usepackage{multirow} 

\usepackage{nicefrac}       
\usepackage{epsfig}
\usepackage{graphicx}
\usepackage{amssymb}
\usepackage{url}
\usepackage{amsthm}
\usepackage{bm}
\usepackage{footmisc}
\usepackage{color}
\usepackage{xfrac}
\usepackage[T1]{fontenc}
\usepackage{pifont}
\usepackage{hyperref}       
\usepackage{flushend} 
\usepackage{units}

\def\R{\mathbb{R}}

\def\l{\left}
\def\r{\right}

\def\piR{\pi_{\tilde{A},\hat{w}}^{R}}
\def\piE{\pi_{\tilde{A},\hat{w}}^{E}}

\def\piRd{\tilde{\pi}_{\tilde{A},\hat{w}}^{R}}
\def\piEd{\tilde{\pi}_{\tilde{A},\hat{w}}^{E}}

\def\({\l(}
\def\){\r)}
\def\[{\l[}
\def\]{\r]}

\begin{document}

\title{Image-to-Image Translation Framework \\ Embedded with Rotation Symmetry Priors}

\author{Feiyu Tan, Heran Yang, Qihong Duan, Kai Ye, Qi Xie, and Deyu Meng
\thanks{Feiyu Tan, Heran Yang, Qihong Duan, Qi Xie, and Deyu Meng are with the School of Mathematics and Statistics, Xi'an Jiaotong University, Shaanxi, P.R.China (e-mail: tanfy929@stu.xjtu.edu.cn, hryang@mail.xjtu.edu.cn, khtuan@126.com, xie.qi@ mail.xjtu.edu.cn, dymeng@mail.xjtu.edu.cn).}
\thanks{Kai Ye is with MOE Key Lab for Intelligent Networks and Networks Security, Faculty of Electronic and Information Engineering, Xi’an Jiaotong University, Shaanxi, P.R.China (e-mail: Kaiye@xjtu.edu.cn).}
}

\markboth{Journal of \LaTeX\ Class Files,~Vol.~14, No.~8, August~2021}%
{Shell \MakeLowercase{\textit{et al.}}: Image-to-Image Translation Framework Embedded with rotation symmetry Priors}


\maketitle

\begin{abstract}
Image-to-image translation (I2I) is a fundamental task in computer vision, focused on mapping an input image from a source domain to a corresponding image in a target domain while preserving domain-invariant features and adapting domain-specific attributes. Despite the remarkable success of deep learning-based I2I approaches, the lack of paired data and unsupervised learning framework still hinder their effectiveness. In this work, we address the challenge by incorporating transformation symmetry priors into image-to-image translation networks. Specifically, we introduce rotation group equivariant convolutions to achieve rotation equivariant I2I framework—a novel contribution, to the best of our knowledge, along this research direction. This design ensures the preservation of rotation symmetry, one of the most intrinsic and domain-invariant properties of natural and scientific images, throughout the network. Furthermore, we conduct a systematic study on image symmetry priors on real dataset and propose a novel transformation learnable equivariant convolutions (TL-Conv) that adaptively learns transformation groups, enhancing symmetry preservation across diverse datasets. We also provide a theoretical analysis of the equivariance error of TL-Conv, proving that it maintains exact equivariance in continuous domains and provide a bound for the error in discrete cases. Through extensive experiments across a range of I2I tasks, we validate the effectiveness and superior performance of our approach, highlighting the potential of equivariant networks in enhancing generation quality and its broad applicability. 
Our code is available at \href{https://github.com/tanfy929/Equivariant-I2I}{https://github.com/tanfy929/Equivariant-I2I}.

\end{abstract}


\newtheorem{Thm}{Theorem}
\newtheorem{Rem}{Remark}
\newtheorem{Lemma}{Lemma}
\newtheorem{Cor}{Corollary}

\section{Introduction}
\IEEEPARstart{I}{mage}-to-image translation (I2I) is a core research problem in computer vision, concerned with learning a mapping that converts an input image from a source domain into a corresponding image in a target domain \cite{isola2017image, chen2022vector}. The primary objective is to preserve domain-invariant features while effectively adapting domain-specific attributes such as color schemes, textural patterns, and brushstrokes. 
This enables the generation of new images based on certain conditions. 
Typical tasks, such as image synthesis \cite{wang2023quantitative}, style transfer  \cite{deng2022stytr2}, and image enhancement \cite{sun2023umgan}, have all achieved remarkable success through the I2I framework.

In the past few years, deep learning-based methods have made significant advancements across various applications,  particularly including the I2I task \cite{albahar2019guided, zhang2020cross, zhou2021cocosnet, shaham2021spatially}. 
It was believed that the success of these methods is mainly due to their data-driven nature, with large-scale and high-quality training datasets playing
a crucial role. By utilizing these comprehensive training datasets, modern deep learning models can effectively learn robust
mappings between different image domains and produce increasingly impressive results.

Unfortunately, in I2I tasks, the high cost for obtaining paired data inevitably leads to the insufficiency of training data, making the learning problem always highly ill-posed and significantly affecting the performance of deep learning-based I2I methods. 
Specifically, current I2I methods largely rely on unsupervised frameworks \cite{zhu2017unpaired, park2020contrastive, xie2023unpaired}, where both the domain-invariant features and domain-specific attributes are hard to be learned and decoupled. Consequently, both the reliability and generalization capability of these methods often hardly satisfy the requirements of complex real-world applications.

It is now widely recognized that embedding prior knowledge into network architectures can effectively mitigate the ill-posed nature of learning problems, especially by alleviating the challenges posed by limited supervised data, and can significantly enhance the performance of deep learning–based image processing methods \cite{kang2024deep, xie2020mhf, liu2024infrared, tan2025ds}.
A typical example is convolutional neural networks (CNNs), which inherently exhibit translation equivariance\footnote{The transformation equivariance in a network means that applying the transformation to the input results in a predictable and consistent change in the intermediate feature maps and output while preserving certain inherent
properties of the system.} by embedding the prior knowledge of image translation symmetry. 
This inherent prior knowledge enables CNNs to outperform fully connected neural networks, establishing them as the dominant approach for image-related tasks. Moreover, recent researches have also verified that embedding more image priors, such as sparse prior \cite{kang2024deep}, low-rankness prior \cite{xie2020mhf}, local smoothness prior \cite{liu2024infrared}, and clusterable prior \cite{tan2025ds}, into the network structure can significantly improve the network's performance in various unsupervised or data-limited cases.
Exploring and embedding more fundamental image priors into network architectures is therefore expected to play a significant role in improving I2I tasks that suffer from their highly ill-posed issue.

\begin{figure}
    \centering
    \includegraphics[width=1\linewidth]{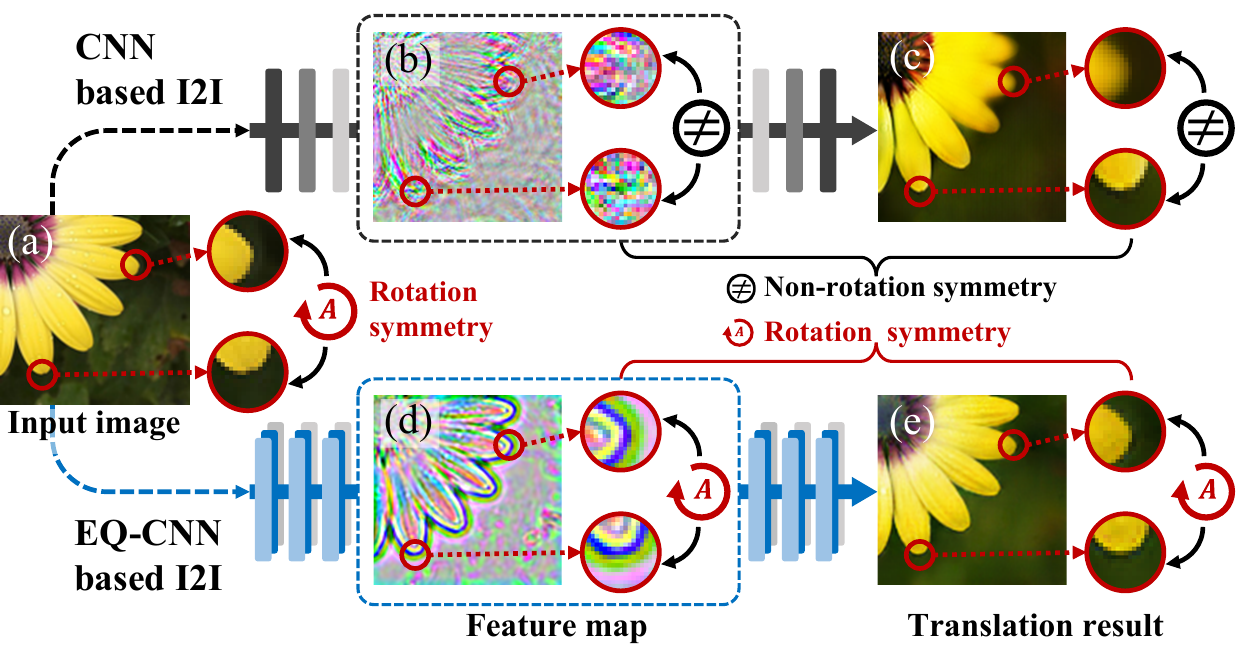}
    \caption{    
    (a) An input image for I2I containing numerous rotationally symmetric structures (illustrated using the iphone2dslr example).
(b–c) Intermediate feature maps and translation results from a standard CNN, showing noticeable symmetry loss and structural artifacts.
(d–e) Intermediate feature maps and translation results from the rotation equivariant network, demonstrating superior preservation of feature symmetry and producing higher-fidelity outputs.
    }
    \label{Fig1}\vspace{-3mm}
\end{figure}

In particular, embedding symmetry priors under various transformations, commonly exhibited in natural and scientific images, into I2I networks holds promise not only for alleviating the ill-posed nature of the task, but also for enabling the network to be equivariant with respect to those transformations. Notably, local feature symmetry under appropriate transformations is an intrinsic domain-invariant property in I2I tasks, as illustrated in Fig. \ref{Fig1}(d) and (e), and should be preserved both before and after the translation. This highlights that achieving equivariance to proper transformations is well aligned with the core principle of domain-invariant feature preservation in I2I learning.

However, to the best of our knowledge, existing I2I methods are mainly built upon traditional convolutional networks, while lack the capability to embed symmetry prior under complex transformations.
For instance, as shown in Fig \ref{Fig1}(a), rotation symmetry of local image patterns is one of the most typical image priors beyond translation symmetry.  
Nevertheless, as illustrated in Fig. \ref{Fig1}(b), since standard CNN is not rotation equivariant, 
its extracted features tend to exhibit chaotic representations and struggle to preserve the inherent rotation symmetry of the input.
Notably, as illustrated in Fig. \ref{Fig1}(c), identical local structures in varying orientations in the input image may suffer from structural inconsistency after being processed by standard CNN-based I2I networks, which violates the domain-invariant feature preservation principle of I2I.

To release the aforementioned issues, we focus on exploring more transformation symmetry prior of image datasets
and designing proper transformation equivariant I2I frameworks.
Specifically, the contribution of this work can be mainly summarized as follows:

1) By adopting rotation group equivariant convolutional networks (EQ-CNN) \cite{cohen2016group, xie2025rotation} into the I2I frameworks, we achieve rotation equivariant I2I for the first time. This effectively preserves rotation symmetry prior, one of the most critical domain-invariant features, through the entire network flow.
As shown in Fig. \ref{Fig1}(d), for rotation symmetric contents, the proposed EQ-CNN-based I2I ensures that all feature maps exhibit symmetry similar to that of the input image. This property naturally leads to rotation symmetry in the final I2I output, as demonstrated in Fig. \ref{Fig1}(e). Consequently, the method effectively overcomes the aforementioned limitation of CNN-based I2I methods (illustrated in Fig. \ref{Fig1}(b) and (c)), which often fail to maintain rotation symmetric structures. This enhances the generation capability and overall performance.

2) We design a transformation learnable equivariant convolution (TL-Conv)  to further preserve the prior of symmetry with respect to an adaptively learned transformation group. This further enhances the proposed equivariant I2I by improving its adaptation to diverse datasets where local image structures exhibit stronger symmetry than those under strict rotation transformations.
Specifically, we first demonstrate the necessity of transformation learnable equivariant network by three steps: 
(I) proposing a well-designed learnable transformation group; 
(II) theoretically studying the relationship between the symmetry of datasets and the equivariance of networks;
(III) verifying the better symmetry with respect to the proposed group compared to the strict rotation transformation group through numerical experiments on commonly used image datasets. 
Next, we redesign the group equivariant convolution framework by developing transformation learnable convolution filters aligned with our proposed transformation learnable group, and adopting them to the EQ-CNN framework. To the best of our knowledge, this should be the first equivariant convolution method capable of adaptively learning transformation groups over a broad spectrum, which fits I2I tasks better compared to conventional CNNs or EQ-CNNs.

3) We provide a comprehensive theoretical analysis of the equivariance error for the proposed method. First, we prove that the equivariance of every convolution layer, including the input, intermediate, and output layers, as well as the entire network, is exact in the continuous domain, even when the  transformation group dynamically varies within the designed space. We then show that the equivariance becomes approximate when the convolution filters are discretized and the symmetry group is learned adaptively during training. Nevertheless, we can still derive a theoretical bound for the equivariance error in this complex case. Our theoretical results first time show that only when the correct transformation is learned for the dataset, the equivariance error can approach zero as the image mesh size diminishes. It should be noted that this fact has often been overlooked in previous work, as prior studies frequently assumed that data symmetry is tied strictly to rotation transformations, neglecting the specificity of different datasets.

4) Experiments on typical image-to-image translation tasks are implemented to validate the effectiveness of our method, including natural image conversion, face rejuvenation, and multi-contrast medical image translation. The results demonstrate its high efficiency and superior performance. Additionally, we further confirm the advantages of our approach through experiments on super-resolution, denoising, and rain removal, highlighting its broad potential and applicability across a range of computer vision tasks.

The remainder of this paper is organized as follows. Sec. \ref{Related Work} provides an overview of related work. Sec. \ref{Rot I2I} presents the prior knowledge of rotation equivariant convolutions and the proposed I2I framework. Sec. \ref{TLEConv} introduces the proposed transformation learnable equivariant convolution, and also provides the equivariance error analysis of our method. Sec. \ref{experiment} demonstrates experimental results in evaluating the performance of the equivariant methods. The paper finally concludes with a future work discussion in Sec. \ref{conclusion}.

\section{Related Work}\label{Related Work}

\subsection{Image-to-Image Translation}

Image-to-image translation (I2I) is a key task in computer vision that focuses on translating an input image from one domain to an output image in another domain, while maintaining essential structural and semantic information \cite{isola2017image, pang2021image}. 
I2I has gained significant attention due to its wide range of applications, including image synthesis \cite{huang2018introduction}, style transfer \cite{jing2019neural}, super-resolution \cite{kaji2019overview}, image denoising \cite{tian2020deep}, and image inpainting \cite{zhao2020uctgan}. Isola et al. \cite{isola2017image} first applied conditional GAN to an I2I problem by proposing pix2pix to solve a wide range of supervised I2I tasks \cite{pang2021image}. Similarly, there has been a great deal of subsequent research on supervised image-to-image translation, which effectively leverages the information from paired datasets \cite{wang2018discriminative, zhang2020cross, zhou2021cocosnet}.

However, obtaining a large amount of paired data requires significant resource costs, making it common to address image-to-image translation as an unsupervised task using unpaired data. This approach, which involves learning a mapping between two image domains without paired examples for training, is particularly valuable in scenarios where acquiring paired data is challenging, such as in artistic style transfer \cite{kotovenko2019content, chen2021dualast}, medical imaging \cite{li2023multi, meng2024multi}, and other cross-domain image synthesis \cite{taigman2016unsupervised, zhou2022hrinversion}. 
Techniques like CycleGAN and other adversarial networks have been successfully applied to this problem by leveraging cycle consistency loss \cite{zhu2017unpaired, lira2020ganhopper, li2018unsupervised}, which ensures that an image transformed from one domain to the other can be reverted to its original form. 

While the cycle consistency constraint can certainly eliminate the dependence on paired data, it tends to force the model to generate a translated image that preserves all input information for reconstruction \cite{pang2021image}. To enhance the efficiency and performance of I2I methods, contrastive learning has been widely adopted \cite{chen2020simple, he2020momentum}. They suggest that this encourages corresponding image patches to map to similar points in the learned feature space while pushing dissimilar patches (negative samples) farther apart \cite{park2020contrastive, wang2021instance}. 
In recent years, considerable effort has been made in the careful design of network architectures to further extract image features, construct a coherent latent space, and decouple the content and style features. For example, CoMoGAN \cite{pizzati2021comogan} introduced a new functional instance normalization layer and residual mechanism, which together disentangle image content from position on the target manifold. Puff-Net \cite{zheng2024puff} designed a novel transformer model that includes only the encoder. A content feature extractor and a style feature extractor are constructed, based on which pure content and style images can be fed to the transformer. Similarly, VQI2I \cite{chen2022vector} and QuantArt \cite{huang2023quantart} introduced vector quantization technique \cite{esser2021taming} in order to construct a sharing feature space and further enable the ability of image extension with flexibility in both intra and inter domains.

Although current I2I methods have achieved promising results, they have yet to incorporate the transformation symmetry priors for releasing the ill-pose problem  and better domain-invariant feature
preservation. This work thus investigates the integration of rotation symmetry priors into I2I architectures.

\subsection{Equivariant CNNs}

Equivariance of a mapping refers to the property that a transformation applied to the input results in a predictable and consistent transformation of the output \cite{lenc2015understanding, smidt2021euclidean}. Equivariant convolutional neural networks are a class of neural networks designed to preserve specific symmetries or transformations in data \cite{cohen2021equivariant}. In earlier approaches, data augmentation \cite{krizhevsky2012imagenet} is a widely used technique to simulate transformations, such as rotations, translations, and flips in the training set. Its concept is to enhance the training dataset by adding transformed versions of the original samples, thereby helping the model become more robust to various transformations \cite{quiroga2020revisiting}. However, it lacks formal theoretical guarantees, meaning that the model may not consistently preserve equivariance across all types of transformations. Furthermore, relying on data augmentation can lead to inconsistent performance, as the model may not always handle transformation symmetries in a predictable or reliable way \cite{xie2022fourier}.

Subsequent works have explored directly integrating rotation equivariance into the deep network architecture. G-CNN \cite{cohen2016group} first constructed an equivariant network, using G-convolutions that enjoy a substantially higher degree of weight sharing than regular convolution layers, increasing the expressive capacity of the network without increasing the number of parameters. HexaConv \cite{hoogeboom2018hexaconv} showed how to efficiently implement planar convolution and group convolution over hexagonal lattices, by reusing existing highly optimized convolution routines. In order to enrich more symmetries in networks, lots of research have been proposed, such as ORNs \cite{zhou2017oriented}, RotEqNet \cite{marcos2017rotation}, H-Nets \cite{worrall2017harmonic}, and so on. 
Later, filter parameterization techniques have been exploited to construct equivariant CNNs. For example, SFCNNs \cite{weiler2018learning} and E2-CNN \cite{weiler2019general} employed steerable filters to efficiently compute orientation-dependent responses for many orientations and achieve arbitrary degree rotation equivariance. Subsequently, PDO-eConvs \cite{shen2020pdo} and PDO-eS2CNNs \cite{shen2021pdo} discretized the system using the numerical schemes of PDOs, deriving approximately equivariant convolutions and analyzing the equivariance error. F-Conv \cite{xie2022fourier} proposed a filter parametrization method based on Fourier series expansion and constructed a high-accuracy equivariant convolution. Li et al. \cite{li2024affine} further built affine equivariant networks based on differential invariants from the viewpoint of symmetric PDEs, without discretizing or sampling the group, and validated the performance across classification tasks. 
Recently, Xie et al. \cite{xie2025rotation} proposed a bicubic basis-based filtering parametrization for rotation equivariant convolution, named B-Conv, which has been demonstrated to be more natural than previous methods.

Although  equivariant convolution approaches have achieved perfect  rotation symmetry prior embedding, the datasets encountered in image-to-image translation are highly varied, and the symmetry of local structures within images tends to be more complex. Equivariant convolutional networks with fixed transformations may no longer be suitable for all tasks. To address this issue, this study further explores transformation learnable equivariant networks that can adapt to different datasets and learn unknown transformation patterns.



\section{Rotation Equivariant I2I Methods}\label{Rot I2I}

In this section, we first introduce the preliminary knowledge of equivariant convolutions \cite{xie2022fourier}, and then provide the construction scheme for rotation equivariant I2I frameworks based on these operations.

\subsection{Preliminaries of Rotation Equivariant Convolutions}

\textbf{Group Equivariance:}
Equivariance of a transformation means that a transformation on the input will result in a predictable transformation on the output. 
Formally, let $\tilde{\Psi}$ denote a mapping from the input space to the output space, and $G$  is a group of transformations. Then, following the definition in previous works \cite{weiler2019general, shen2020pdo, xie2022fourier}, 
$\tilde{\Psi}$ is equivariant with respect to $G$ if for any $g \in G$, 
\begin{equation}\label{equivariance}
    \tilde{\Psi}[\pi_g^{R}](I) = \pi_g^{E}[\tilde{\Psi}](I),
\end{equation}
where $I$ represents the input image, $\pi_g^{R}$ and $\pi_g^{E}$ denote how the transformation $g$ acts on input image and output features respectively,  $[~\!\cdot\!~]$ denotes the composition of functions. 

\begin{figure*}[t]
    \centering\vspace{-3mm}
    \includegraphics[width=1\linewidth]{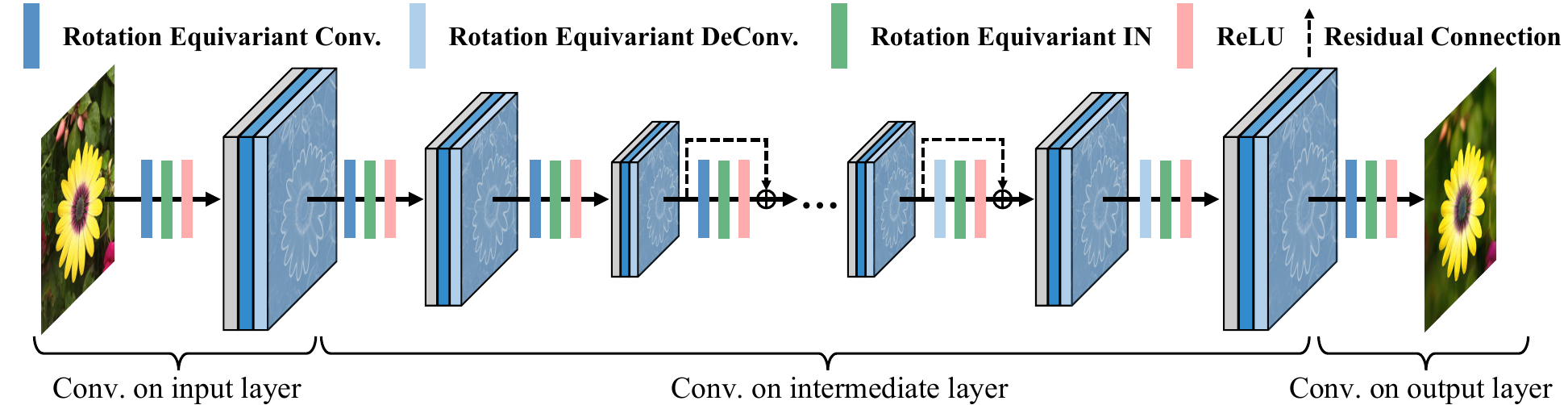}
    \caption{An example of rotation equivariant network architecture, which consists of equivariant convolution, equivariant transposed convolution (DeConv.), Instance Normalization (IN), and activation functions (ReLU).}
    \label{UNet}
\end{figure*}

\textbf{Rotation Group and Feature Map:} In this work, we follow the framework of a typical variant \cite{xie2022fourier} of the classical group convolution to construct equivariant I2I models, which achieves state-of-the-art (SOTA) performance in low-level image processing tasks. For simplicity, we only introduce the definition of single channel convolutions, which can be naturally extended to multi channels and transposed convolutions. 

Formally, we denote the input image as $I\in \R^{n\times n}$, where $n$ is the height and width of the image. Meanwhile, the feature map of the corresponding convolutional network is denoted as $F$. It should be noted that, different from the two-dimensional structure of feature maps in traditional convolutional networks, the feature maps in equivariant convolutional networks are all three-dimensional, i.e., $F \in \R^{n\times n \times t}$, where $t$ is the number of elements in the discrete equivariance transformation group $S$. In this section, we have 
 \begin{equation}\label{S-Group}
S\!=\!\l\{\!A\! \!=\!\!
\begin{bmatrix}
    \cos \theta\! & \!\sin \theta \\
   -\sin \theta\! &  \!\cos \theta
  \end{bmatrix}\bigg| \theta \!=\! \frac{2\pi k}{t}, k = 0,1,\!\cdots\!,t\!-\!1
\r\},
\end{equation}
where the element $A$ is a rotation transformation matrix, which is also used as the index of the group dimension of $F$ and its corresponding filters.


\textbf{Equivariant Convolution Filters:} It is should be noted that the filters in EQ-CNNs are also in a different structure compared to those in traditional CNNs. Particularly, in the EQ-CNN framework, the filters for input, intermediate, and output layers are distinct from each other. The filters in the input and output layers are 3D tensors with a shape of $p\times p \times t$, where $p$ denotes the filter size. In contrast, the filters in the intermediate layer are in a 4D structure of shape $p\times p\times t\times t$. 
Furthermore, in rotation equivariant CNNs,  the filters must rotate along the third dimension (group dimension) while maintaining their structures, and filter parametrization thus becomes necessary. 
Specifically, we denote the filters in the input, intermediate, and output layers as 
$\tilde{\Psi} = \l[\tilde{\Psi}_{ij}^A\r]$, $\tilde{\Phi} = [\tilde{\Phi}_{ij}^{A,B}]$, and $\tilde{\Upsilon} = \l[\tilde{\Upsilon}_{ij}^B\r]$, where  $i, j = 1,2, \cdots, p$, $A,B\in S$. Then by adopting the filter parametrization technique (which will be described in the next paragraphs), the elements in filters for input, intermediate, and output layers can be respectively defined as follows:
\begin{equation}\label{Phi_}
	\begin{split}
		&\tilde{\Psi}_{ij}^A = \varphi_{in}\left(A^{-1}\delta_{ij}\right),\\
		&\tilde{\Phi}_{ij}^{B,A} = \varphi_A\left(B^{-1}\delta_{ij}\right),\\
		&\tilde{\Upsilon}_{ij}^{B} = \varphi_{out}\left(B^{-1}\delta_{ij}\right),
	\end{split}
\end{equation}
where $\varphi_{in}$, $\varphi_A$ and $\varphi_{out}$ are parameterized filters, $\delta_{ij} = \left(\left(i-\nicefrac{(p+1)}{2}\right)h, \left(j-\nicefrac{(p+1)}{2}\right)h\right)^T$ is the mesh grid for obtaining discretized filters, and $h$ is the mesh size.

\textbf{Filter Parametrization:} Following the previous works \cite{weiler2018learning, shen2020pdo, xie2022fourier}, the functional filter $\varphi: \mathbb{R}^2\to\mathbb{R}$ can be expressed as the linear combination of a set of basis functions $\{\varphi_k\}_{k=1}^{K}$: 
\begin{equation}\label{Parametrization}
  \varphi(\delta)=\sum_{k=1}^{K} v_k\varphi_k(\delta),
\end{equation}
where $K$ is the number of basis functions, $v_k$ is the $k$-th learnable coefficient. Based on \eqref{Parametrization}, the filters in (\ref{Phi_}) can be effectively represented and learned. 

\textbf{Calculation of Convolutions:} According to the EQ-CNN framework in \cite{xie2022fourier}, the convolution of input layer can be calculated by performing commonly used convolution operation along the group dimension, i.e.,  
\begin{equation}\label{input_}
\left(\tilde{\Psi}\star I\right)^A =\tilde{\Psi}^A*I,~ \forall A\in S,
\end{equation}
where $*$ is the commonly used discrete 2D convolution.
Besides,  the intermediate convolution  can be calculated by:
\begin{equation}\label{regular_}
  \left(\tilde{\Phi}\star F\right)^B =\sum_{A\in S}\tilde{\Phi}^{B,B^{-1}A}*F^{A},~  \forall B\in S,
\end{equation}
and the convolution of output layer can be calculated by:
\begin{equation}\label{output_}
  \tilde{\Upsilon}\star F =\sum_{B\in S}\tilde{\Upsilon}^{B}\ast F^{B},~ \forall A\in S.
\end{equation}

Based on the above rotation equivariant convolutions, it is easy to deduce rotation equivariant transposed convolution for  upsampling\cite{weiler2018learning}. Then, rotation equivariant networks can be built, and it has been verified that this architecture inherently incorporates the rotation symmetry prior, thereby yielding enhanced output quality \cite{weiler2018learning, shen2020pdo, xie2022fourier}.

\subsection{Rotation Equivariant I2I Framework}
\textbf{Rotation Equivariant Framework Design:}
For I2I tasks, the network architectures typically comprise not only convolutional layers but also activation functions, downsampling and upsampling operations along with normalization layers to ensure proper feature extraction. 
Fortunately, activation functions as well as commonly used downsampling and upsampling operators have been proven to be rotation equivariant \cite{liu2025rotation}. In addition, in previous works, equivariant ameliorations have also been proposed for normalization layers such as Instance Normalization (IN) and Batch Normalization (BN). 
Fig. \ref{UNet} shows a typical example for easy understanding.

\begin{figure*}
    \centering
    \includegraphics[width=0.9\linewidth]{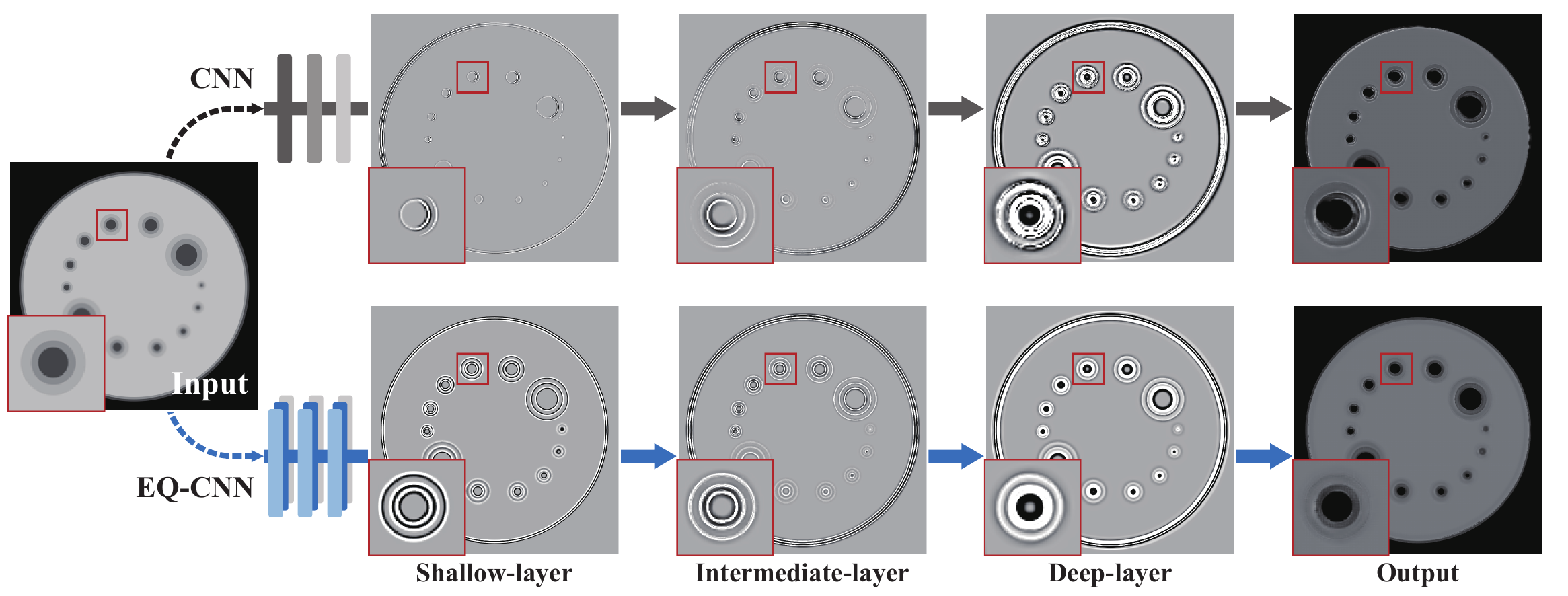}
    \caption{The feature maps of the trained I2I CNN and EQ-CNN, respectively, along with their corresponding outputs. The two exploited networks are both well trained for MRI translation (details can be found in Sec. \ref{experiment}-B)
    We display the feature maps from shallow to deep layers, where a single channel is arbitrarily selected for representing each layer.
    }
    \label{cartoon_feature}
\end{figure*}

It is important to note that in the EQ-CNN framework, all the filters are rotated by degrees of $\nicefrac{2\pi k}{t}, (k=0,1,...,t-1)$ and reused, indicating that the parameters of the parameterized filters are reused for $t$ times. Therefore, with the same number of feature channels, the rotation equivariant framework has fewer parameters, specifically, $\nicefrac{1}{t}$ of the corresponding CNN. Based on the relationship between the number of parameters and the generalization ability of the model, it can be deduced that a lower-parameter model has better generalization ability \cite{elesedy2022group}.

\textbf{Visualization and Verification of Equivariance:}
For verification of the equivariance of the proposed I2I framework, we illustrate the feature maps and output of the proposed EQ-CNN-based I2I network, compared to the commonly used CNN-based method. 
As shown in Fig. \ref{cartoon_feature}, we select an input image composed of distinct geometric shapes to facilitate clear observation of rotation equivariance. 
The image is processed through well trained I2I models for MRI modality translation\footnote{The details of this can be referred to the experimental Section V-B.}, with feature visualizations displayed from shallow to deep layers alongside their respective outputs. 
Evidently, the conventional CNN-based model struggles to preserve the intrinsic symmetry of the image during feature extraction, whereas the EQ-CNN-based one consistently and effectively captures rotation symmetric structures through the entire network flow. This not only verifies that the proposed EQ-CNN-based I2I framework is indeed rotation equivariant, but also shows that it has a better ability for preserving domain-invariant features.



\section{Transformation Learnable Equivariant Convolutions for I2I}\label{TLEConv}

\subsection{Motivation: From Symmetry to Equivariance}
While rotation equivariant convolutions and their corresponding networks effectively preserve strict rotation symmetry, this symmetry typically exists only in specific datasets, such as overhead-shot remote sensing or pathological images. In contrast, images for I2I tasks often originate from more complex shooting scenarios, requiring the processing of non-overhead images.  Clearly, strict rotation equivariant networks can be ill-suited for such data.

\begin{figure}
    \centering
    \includegraphics[width=\linewidth]{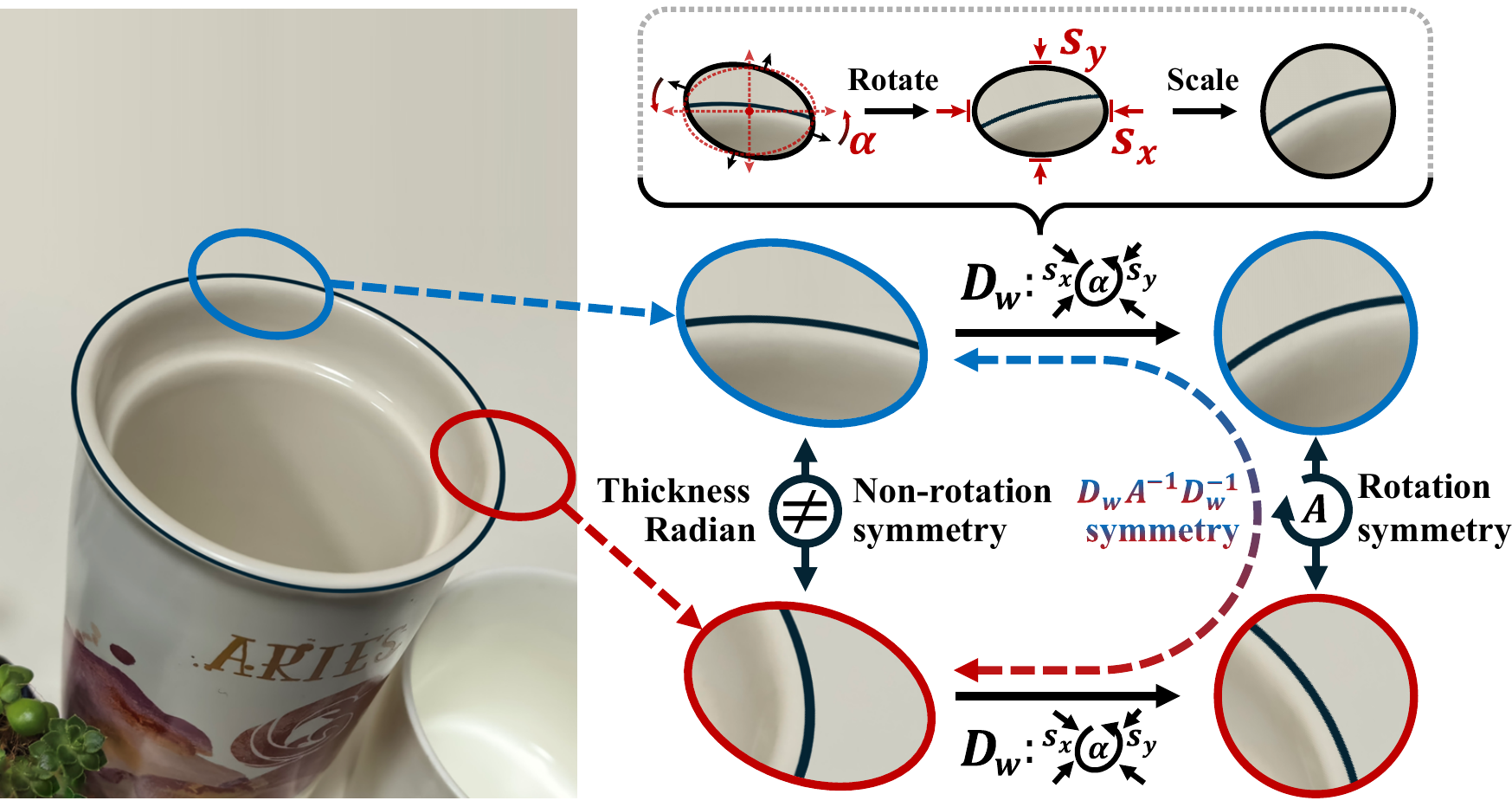}
    \caption{Illustration of the complex symmetry of local structures in real images, where obvious differences in the curvature and thickness of different parts of the cup rim are caused by the non-overhead shooting perspective. This non-strict rotation symmetry can be converted into strict symmetry by additionally adopting a learnable transformation.}
    \label{cup}\vspace{-2mm}
\end{figure}

As the example shown on the left side of Fig. \ref{cup}, two image patches extracted from the rim of the cup clearly exhibit correlated structures. Particularly, the real-world objects corresponding to these two image patches exhibit identical geometric shapes. However, due to the shooting perspective, differences in curvature and thickness arise between them, making it impossible to characterize their symmetry through simple rotation transformations.

Fortunately, this disruption to rotation symmetry caused by the shooting perspective can be mitigated through appropriate affine transformations. As shown in the right side of Fig. \ref{cup}, adopting proper affine operations (rotation and scaling) can well restore the strict rotation symmetry between the image patches.
Building upon this insight, we focus on investigating the rotation group coupled with learnable transformations (denoted as $D_w$ in Fig. \ref{cup}, to be detailed later) and develop corresponding equivariant convolutions.

Specifically, we conduct the following learnable transformation group\footnote{It can be proved that for $\forall w$, $S_w$ is a group. Please refer to the supplementary material for more details.}:
\begin{equation}\label{Sw-group}
    S_w = \{D_w A D_w ^{-1}|A \in S \},
\end{equation}
where $D_w$ is a learnable affine matrix, formally defined as:
\begin{equation}\label{D}
    \begin{split}
        D_w\!=\!\!\begin{bmatrix}
\cos\! \alpha \!&\! \sin \alpha \\
\!-\sin \alpha \!&\! \cos \alpha
\end{bmatrix}\!\begin{bmatrix}
s_{x} \!&\!\! 0 \\
0 \!&\!\! s_y 
\end{bmatrix}
\!\!=\!\!\begin{bmatrix}
s_x \cos \alpha \!&\! s_y \sin \alpha \\
-s_x \sin \alpha \!&\! s_y \cos \alpha
\end{bmatrix}\!,
    \end{split}
\end{equation}
and ${w}=\left[{\alpha}, {s_x}, {s_y}\right]$ denotes the learnable parameters. 

It is noteworthy that the parameters in $w$ carry distinct physical interpretations. As illustrated in Fig. \ref{cup}, $s_x$ and $s_y$ correspond to adaptive scaling coefficients along two orthogonal axes\footnote{We can reasonably assume that they lie within the range of $[0.5, 1.5]$, as real-world affine transformations caused by varying shooting perspective will not lead to significant scaling.}, with $\alpha$ governing the principal direction of adaptive scaling. More importantly, we can also observe that, in non-overhead view images, local image structures potentially exhibit stronger symmetry under transformations from group \eqref{Sw-group}, as compared to strict rotation transformations. 
Particularly, for a given dataset (assuming the images share a uniform camera perspective), there will exist a specific parameter vector\footnote{Detailed explanations and formally definition regarding $\hat{w}$ can be found in the supplementary material.}, i.e., $\hat{w}=\left[\hat{\alpha}, \hat{s_x}, \hat{s_y}\right]$,  such that the dataset achieves best transformation symmetry with respect to the group $S_{\hat{w}}$. 

Specifically, for any rotation matrix $\tilde{A}$ and the aforementioned transformation parameters $\hat{w}$, 
the transformation $g \in S_{\hat{w}}$ acts on an image function $r\in C^\infty(\mathbb{R}^2)$  by:
\begin{equation}\label{new_piR}
	\piR[r](x) = r\left(D_{{w}} \tilde{A}^{-1}D_{{w}}^{-1}x\right), \forall x\in\mathbb{R}^2.
\end{equation}
Then, the symmetry of local image features can be expressed by the following equation: 
\begin{equation}\label{symmetry}
    R\left(\piR[r]\right) = R(r),
\end{equation}
where $R$ denotes the local feature function (which encompasses common image regularization functions).

Prior research has demonstrated that for certain specific tasks, when datasets exhibit intrinsic symmetry, the network operators should maintain corresponding equivariance \cite{celledoni2021equivariant}. This insight prompts a deeper examination of the relationship between symmetry and equivariance. Here, we further prove the following theorem for I2I tasks:
\begin{Thm}\label{I2I_eq}
For the following I2I translation model:
	\begin{equation}
		\operatorname{I2I}(\hat{r}) = \arg \min_{r}\int_{\R^2}\left(\hat{E}[\hat{r}](x)-E[r](x)\right)^2 dx + R(r),
	\end{equation}
where $\hat{r}\in C^\infty(\mathbb{R}^2)$ is the 2D function representing the input image, $E(\cdot)$ denotes the domain-invariant feature extractor, and $R(\cdot)$ denotes a local feature regularization function for I2I. If the local image feature corresponding to $R$ is symmetry with respect to $\piR$,  i.e.,  $\forall \tilde{A}\in S$ and $\hat{w}\in\mathbb{R}^3$, $R(\piR[r]) = R(r)$, then the model $\operatorname{I2I}$ is equivariant, in the sense that
	\begin{equation}
		\operatorname{I2I}[\piR[\hat{r}]] = \piR[\operatorname{I2I}[\hat{r}]].
	\end{equation}
\end{Thm}

Theorem \ref{I2I_eq} highlights the strong connection between the data's symmetry\footnote{We have validated that the image dataset exhibits stronger symmetry under the transformation in \eqref{Sw-group} through extensive experiments, in Sec. V-A.} and the network's equivariance in I2I tasks, i.e., the transformation symmetry in the image data necessarily requires the model architecture to be equivariant correspondingly. 
We thus focus on investigating Transformation-Learnable Equivariant Convolution (TL-Conv), which possesses the potential to preserve better symmetry in data-driven setting frameworks. Specifically, we optimize a parameter $w$ to approximate $\hat{w}$, and enforce the network equivariance under the transformation group $S_w$.

\begin{figure*}[t]
\vspace{0.5mm}
\hspace{-0mm}\includegraphics[width=1.0\linewidth]{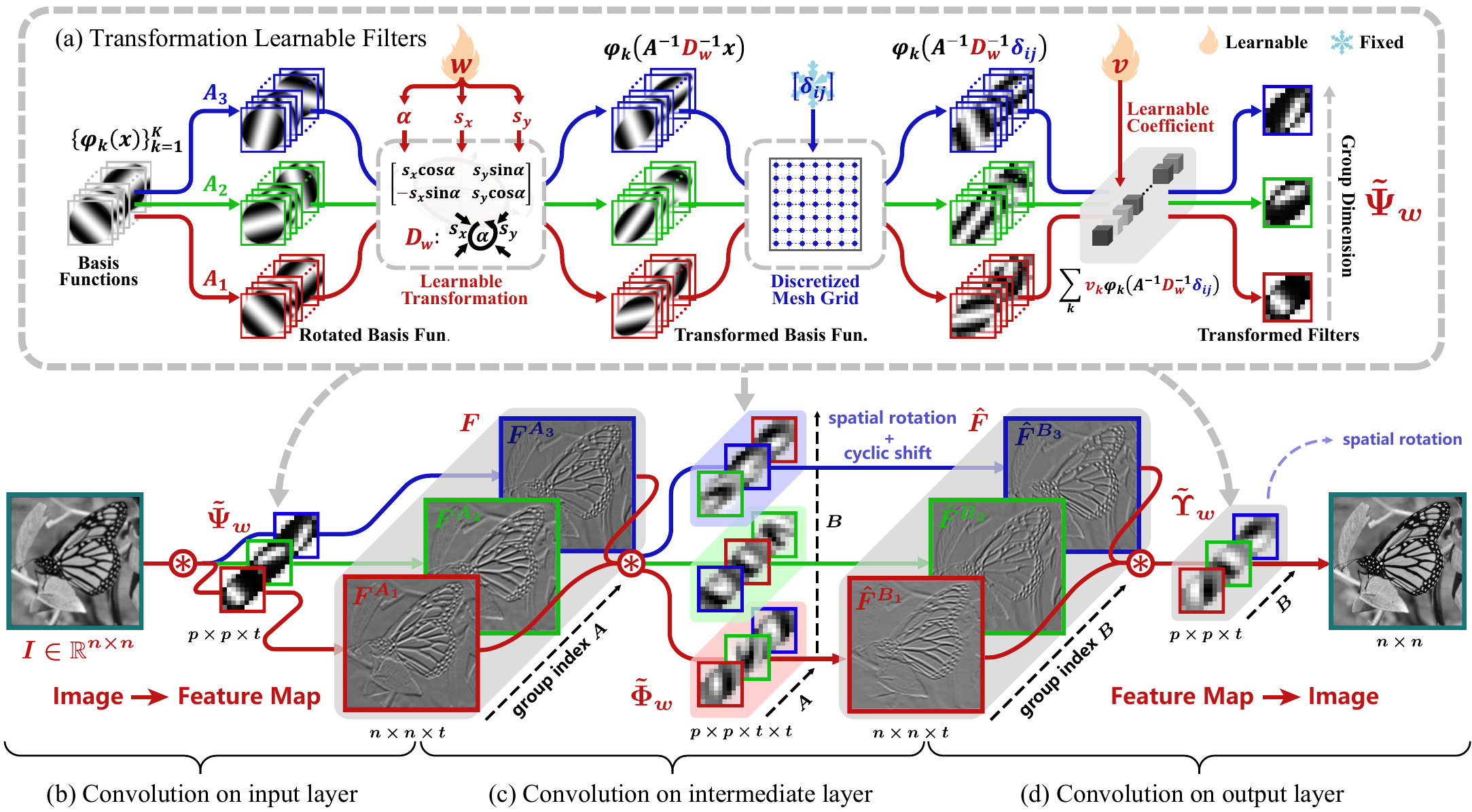}
\vspace{-5mm}
  \caption{Illustration of example network constructed by the proposed transformation learnable equivariant convolutions, where we set transformation group as $\nicefrac{2\pi i}{3}$ rotations, $i = 1,2,3$. (a) The construction process of the proposed learnable filters. A selected set of basis functions undergo rotation and learnable affine transformations. After discretization and the linear combination with coefficients $v$, the filters used in practice are obtained. (b)-(d) Input layer, intermediate layer, and output layer, respectively.}
\label{network}
\vspace{-1.5mm}
\end{figure*}

\subsection{Tool: Transformation Learnable Filters}
\subsubsection*{\bf Filters in the Continuous Domain}
As demonstrated in (\ref{Phi_}), in the equivariant convolution frameworks, the filters must undergo  transformations with respect to the group elements along the group dimension. 
Since the group $S_w$ in \eqref{Sw-group} contains learnable elements, transformation learnable filters are essential for building a transformation learnable equivariant convolution framework. 

In this study, we exploit the  filter parametrization technique (\ref{Parametrization}) for constructing the transformation learnable filters. Formally, as shown in the left part of Fig. \ref{network}(a), in the continuous domain, we express the transformation learnable filter as:
\begin{equation}\label{TL-filter}
  \varphi(A^{-1}D_w^{-1}x)=\sum_{k=1}^{K} v_k\varphi_k(A^{-1}D_w^{-1}x),
\end{equation}
where $A$ is the rotation matrix (i.e., element in $S$), $D_w$ denotes the learnable coordinate transformation matrix defined in \eqref{D} and $\varphi_k(x)$, $k = 1,2, \cdots K$ are $K$ chosen basis functions \cite{xie2022fourier}.  Particularly, both $w = [\alpha, s_x, s_y]$ and $v$ are learnable parameters, where  $w$ controls the learnable transformation and $v$ controls the content of the filter.

It is worth noting that in  (\ref{TL-filter}), both the base $\varphi_k$ and the transformation $D_w$ are composed of functions differentiable in PyTorch, ensuring efficient gradient backpropagation for $w$.  Compared with existing equivariant convolutions, which typically employ fixed dictionary functions and learn only the combination coefficients $v_k$,  our approach jointly optimizes $w$, enabling the  equivariance-adapting convolution.

\subsubsection*{\bf Filters in the Discrete Domain}
The filters in \eqref{TL-filter} are in the continuous domain, while in practical applications, we can only use their discrete form. Similar to the discrete filters defined in \eqref{Phi_}, for $i, j = 1,2, \cdots, p$, $A, B\in S$ and transformation $D_w$ is defined in (\ref{D}), we  define the discrete filters for the input layer as:
\begin{equation}\label{newPsi_}
(\tilde{\Psi}_w)_{ij}^A =\sum_k{v_k^{in}\varphi_k(A^{-1}D^{-1}_w\delta_{ij})}:= \varphi_{in}\left(A^{-1}D_w^{-1}\delta_{ij}\right),
\end{equation}
where $\delta_{ij} = \left(\left(i-\nicefrac{(p+1)}{2}\right)h, \left(j-\nicefrac{(p+1)}{2}\right)h\right)^T$ is the fixed mesh grid, 
$\varphi_{in}$ is the parameterized filter of input layer, and $v^{in}_k$ is its  coefficients. 
Please see Fig. \ref{network}(a) for a more intuitive understanding of the complete construction process of the proposed transformation learnable filters.

Similarly, we define the discrete filters for intermediate and output layers as follows:
\begin{equation}\label{newPhi_}
(\tilde{\Phi}_w)_{ij}^{B,A} = \varphi_A\left(B^{-1}D_w^{-1}\delta_{ij}\right),
\end{equation}
\begin{equation}\label{newUpsilon_}
(\tilde{\Upsilon}_w)_{ij}^{B} = \varphi_{out}\left(B^{-1}D_w^{-1}\delta_{ij}\right),
\end{equation}
where $\varphi_A$ indicates the parameterized filter with respect to the channel of feature map indexed by $A$ in intermediate layer,  and $\varphi_{out}$ is the parameterized filter in the output layer.

\subsection{Method:  TL-Conv Framework}
\subsubsection*{\bf TL-Conv in Continuous Domain}
In the continuous domain, the input image can be modeled as a function defined on $\mathbb{R}^2$, denoted as $r(x)$. Similar to previous approaches, the intermediate feature map can be modeled as a function $e(x,A)$ defined on $E(2)= \mathbb{R}^2 \rtimes O(2)$, where $\rtimes$ is a semidirect-product and $O(2)$ is the orthogonal group. 
Following the group convolution framework \cite{cohen2016group, xie2022fourier}, we can now formulate the proposed TL-Conv.

Let $\Psi_w$ denote the continuous TL-Conv in the input layer, with $w$ being its learnable parameters. It maps an input $r\in C^\infty(\R^2)$ to a feature map defined on $E(2)$. Specifically,   $\forall (x, A)\in E(2)$:
\begin{equation}\label{new_input_Conv}
	\Psi_w[r](x,A) \!=\!\!  \int_{\R^2}\!\varphi_{in}\!\left(A^{-1}\!D_w^{-1} \delta \right)\!r(x-\delta)d\sigma(\delta),
\end{equation}
where $\sigma$ is the  measure on $\R^2$ and $\varphi_{in}$ is the parameterized filter as defined in (\ref{newPsi_}). 

Let $\Phi_w$ denote the convolution in the intermediate layer,
which maps a feature map $e\in C^\infty(E(2))$ to another, then for $\forall (x, B)\in E(2)$:
\begin{equation}\label{new_regular_Conv}
	\begin{split}
		\Phi_w[e](x,\!B) \!\!=\!\!\!
		\int_{\!O(2)}\!\int_{\R^2}\!\!\varphi_{\!A}\!\left(B^{-\!1}D_w^{-\!1}\delta\right)\!e(x\!-\!\delta, \!B\!A)d\sigma(\delta)dv(A),
	\end{split}
\end{equation}
where $v$ is a measure on $O(2)$, $A,B\!\in\! O(2)$ denote transformations in the considered group, and $\varphi_A$ is the parameterized filter as defined in (\ref{newPhi_}).

Let $\Upsilon_w$ denote the convolution in the output layer,
which maps a feature map $e\in C^\infty(E(2))$ to a function defined on $\R^2$, for $\forall x\in \R^2$:
\begin{equation}\label{new_output_Conv}
	\Upsilon_w[e](x) \!\!=\!\!\!  \int_{O(2)}\!\int_{\R^2}\!\!\!\varphi_{out}\!\left(\!B^{-1}\!D_w^{-1}\!\delta\right)\!e(x\!-\!\delta, \!B)d\sigma(\!\delta\!)dv(\!B\!),
\end{equation}
where $B \in O(2)$ and $\varphi_{out}$ is the parameterized filter as defined in (\ref{newUpsilon_}).

In \eqref{new_piR}, we have defined the transformation on an image $r$. Here, 
in order to analyze the rotation equivariance of the proposed method, we now formally define the relevant transformations on feature maps.
Specifically, for any rotation $\tilde{A}$ and transformation parameters $\hat{w}$, let $g = D_{\hat{w}}\tilde{A}^{-1}D_{\hat{w}}^{-1}$, it acts on a feature map $e\in C^\infty(E(2))$ by: 
\begin{equation}\label{new_piE}
	\piE[e](x,A) = e(D_{\hat{w}} \tilde{A}^{-1}D_{\hat{w}}^{-1}x,\tilde{A}^{-1}A), \forall (x,A)\in E(2).
\end{equation}
It should be noted that $\hat{w}$ is the dataset-dependent parameter, where distinct datasets yield different optimal $\hat{w}$ values that introduce stronger symmetry.

It can be deduced that the proposed convolutions are indeed  equivariant with above transformation in the continuous domain. In summary, we have the following conclusion (the detailed proofs are provided in the supplementary material):
\begin{Thm}\label{Thm_c}
For $r\in C^\infty(\R^2)$, $e\in C^\infty(E(2))$, $\forall \tilde{A}\in O(2)$ and $\forall \hat{w}\in \mathbb{R}^3$, when $w = \hat{w}$, the following results are satisfied:
\begin{equation}\label{Thm1}
	\begin{split}
		\Psi_w\left[\piR\left[r\right]\right] &= \piE\left[\Psi_w\left[r\right]\right],\\
		\Phi_w\left[\piE\left[e\right]\right] &= \piE\left[\Phi_w\left[e\right]\right],\\
		\Upsilon_w\left[\piE\left[e\right]\right] &= \piR\left[\Upsilon_w\left[e\right]\right],\\
	\end{split}
\end{equation}
where $\piR$, $\piE$, $\tilde{\Psi}_w$, $\tilde{\Phi}_w$ and $\tilde{\Upsilon}_w$
are defined by (\ref{new_piR}), (\ref{new_piE}), (\ref{new_input_Conv}), (\ref{new_regular_Conv}) and (\ref{new_output_Conv}), respectively.
\end{Thm}

Theorem \ref{Thm_c} demonstrates that even for distinct $\hat{w}$ derived from different datasets, we can optimize $w$ via gradient backpropagation to satisfy $w = \hat{w}$, thereby enforcing equivariance under the corresponding transformation group.

\subsubsection*{\bf TL-Conv in Discrete Domain}
Based on the continuous convolutions above, we then show how to use the discrete filters defined in (\ref{newPhi_}) to construct the discrete convolutions.
We assume that an image $I\in \R^{n\times n}$ represents a two-dimensional grid function obtained by discretizing a smooth function, i.e., for $i, j = 1,2, \cdots, n$,
\begin{equation}\label{I}
	I_{ij} = r(x_{ij}),
\end{equation}
where $x_{ij} = \left(\left(i-\frac{n+1}{2}\right)h, \left(j-\frac{n+1}{2}\right)h\right)^T$ and $h$ is the mesh size. $F$ represents a three-dimensional grid function sampled from a smooth function $e: \R^2\times S\to \R$, i.e., for $i, j = 1,2, \cdots, n$,
\begin{equation}\label{F}
	F_{ij}^A = e(x_{ij}, A),
\end{equation}
where $A\in S$.
Accordingly, we can discretize the continuous convolutions for the input layer, intermediate layer, and output layer (i.e.,(\ref{new_input_Conv}), (\ref{new_regular_Conv}), and (\ref{new_output_Conv})). For $\forall A\in S, \forall w\in \mathbb{R}^3$ and $i,j = 1,2,\cdots,n,$ $\tilde{\Psi}_w$ denotes the convolution of input layer, which can be calculated by:
\begin{equation}\label{new_input_}
\left(\tilde{\Psi}_w\star I\right)^{A}  =(\tilde{\Psi}_w)^A*I, ~ \forall A\in S.
\end{equation}
An example of the input layer is shown in Fig. \ref{network}(b) for easy understanding.

Similarly, for any $B\in S$ and $i,j = 1,2,\cdots,n,$ we can define the discrete TL-Conv of intermediate layer as the discretization of \eqref{new_regular_Conv}:
\begin{equation}\label{new_regular_}
  \left(\tilde{\Phi}_w\star F\right)^B =\sum_{A\in S}\left(\tilde{\Phi}_w\right)^{B,A}*F^{BA},~  \forall B\in S.
\end{equation}
To ensure efficient implementation, we avoid channel transformations on feature maps and instead use an equivalent form of the above equation:
\begin{equation}\label{new_regular_2}
  \left(\tilde{\Phi}_w\star F\right)^B =\sum_{A\in S}\left(\tilde{\Phi}_w\right)^{B,B^{-1}A}*F^{A},~  \forall B\in S.
\end{equation}
It is clear that (\ref{new_regular_2}) can be implemented by applying the commonly used convolution on $F$ and a ${p\times p\times t\times t}$ filter rotated spatially and shifted cyclically along the group dimension, as shown in Fig. \ref{network}(c). 

For $i,j = 1,2,\cdots,n,$ $\tilde{\Upsilon}_w$ denotes the convolution of output layer:
\begin{equation}\label{new_output_}
  \tilde{\Upsilon}_w\star F =\sum_{B\in S}\left(\tilde{\Upsilon}_w\right)^{B}\ast F^{B},~ \forall A\in S.
\end{equation}
A more intuitive illustration of this can be found in Fig. \ref{network}(d).

\subsubsection*{\bf Implementation Details}
The proposed method can be seamlessly integrated into existing deep networks in a plug-and-play manner.
Specifically, by replacing all convolution operators in the CNN-based I2I networks with the proposed ones, we can obtain a transformation learnable equivariant network without altering other architectures of the backbone or the loss functions. 
For the basis functions selection for the filter parametrization framework \eqref{TL-filter}, we refer to the work of B-Conv \cite{xie2025rotation} and choose the bicubic function, which has been validated for its effectiveness in several computer vision tasks. 

Notably, the proposed TL-Conv contains learnable affine transformation parameters $w$, while the intrinsic transformation $\hat{w}$ of the dataset is actually unknown. In this study, we simply set $w$ as a learnable parameter and simultaneously optimize it together with parameter $v$ and other network parameters based on the task's loss function during network training. Although there is no theoretical guarantee that the accurate $\hat{w}$ can be learned in this manner, we empirically find this approach to be effective for performance improvement. Furthermore, it is worth noting that, by setting the rotation angle $\alpha$ of the learnable transformation as constant $0^\circ$ and the scaling factors $s_x$, $s_y$ as $1$ (this is adopted as our initialization scheme during network training), the proposed method degenerates into a standard rotation equivariant convolution, offering a flexible approach for different scenarios. 

\subsection{Theory: Equivariance Error Analysis}
\subsubsection*{\bf Equivariance Error of Single Layer}
In the previous section, we have demonstrated that the proposed convolutions are strict equivariant with zero error in the continuous domain. However, after discretization, approximation errors inevitably emerge.
Next, we will analyze the equivariance approximation errors of these discretized convolutions. 

First, $\forall i, j = 1,2, \cdots, n,  A, \tilde{A} \in S$, and $\hat{w}\in \mathbb{R}^3 $,
we denote the following transformations on $I$ and $F$:
\begin{equation}\label{new_piD}
	\begin{split}
		& \left(\piRd(I)\right)_{ij} \!\!=\! \piR[r](x_{ij}),\\
		& \left(\piEd(F)\right)_{ij}^{A} \!\!= \!\piE[e](x_{ij},A).
	\end{split}
\end{equation}

Then, we deduce the following theorem (the detailed proofs are provided
in the supplementary material).

\begin{Thm}\label{Thm_d}
	Assume that an image $I\in \R^{n\times n}$ is discretized from the smooth function $r:\R^2\to\R$ by (\ref{I}), a feature map $F\in \R^{n\times n \times t}$ is discretized from the smooth function $e:\R^2\times S\to\R$ by (\ref{F}), $|S|=t$, filters $\tilde{\Psi}$, $\tilde{\Phi}$ and $\tilde{\Upsilon}$ are generated from $\varphi_{in}$, $\varphi_{A}, \forall A\in S$ and $\varphi_{out}$ by (\ref{newPsi_}), (\ref{newPhi_}) and (\ref{newUpsilon_}) respectively. If for any $A\in S, ~w\in \mathbb{R}^3, ~ x\in \R^2$, the following conditions are satisfied:
	\begin{equation}\label{condition_t2_}
		\begin{split}
			&|r(x)| , |e(x,A)|\leq F_1,\\
			&\|\nabla r(x)\| , \|\nabla e(x,A) \|\leq G_1,\\
			&\|\nabla^2 r(x) \| , \|\nabla^2 e(x,A) \|\leq H_1,\\
			&|\varphi_{in}(x)|, |\varphi_{A}(x)|, |\varphi_{out}(x)|\leq F_2, \\
			&\|\nabla \varphi_{in}(x) \|, \|\nabla \varphi_{A}(x) \|, \|\nabla \varphi_{out}(x) \|\leq G_2, \\
			&\|\nabla^2 \varphi_{in}(x)\|, \|\nabla^2 \varphi_{A}(x) \| , \|\nabla^2 \varphi_{out}(x) \|\leq H_2, \\
			&\forall \|x\|\geq\nicefrac{(p+1)h}{2},~ \varphi_{in}(x), \varphi_{A}(x), \varphi_{out}(x)  =0,
		\end{split}
	\end{equation}
	where $p$ is the filter size, $h$ is the mesh size, $\nabla$ and ${\nabla}^2$ denote the operators of gradient and Hessian matrix, respectively, then for any $\tilde{A}\in S, w,\hat{w}\in \mathbb{R}^3$, the following results are satisfied:
    \begin{equation}\label{Thm2_}
		\begin{split}
			&\left\|\!\tilde{\Psi}_w\!\star\!\piRd\!\!\left(I\right) \!-\! \piEd\!\!\left(\!\tilde{\Psi}_w\!\star\! I \!\right)\!\right\|_{\infty}
			\!\leq \tilde{C}\!\left\|w\!-\!\hat{w}\right\| \!+\! Ch^2\!, \\
			&\left\|\!\tilde{\Phi}_w\!\star\!\piEd\!\!\left(F\right) \!\!-\! \piEd\!\!\left(\!\tilde{\Phi}_w\!\star\! F\!\right)\!\right\|_{\infty}
			\!\!\leq \!\tilde{C}\!\left\|w\!-\!\hat{w}\right\|t \!+\! Ch^2t,\\
			&\left\|\!\tilde{\Upsilon}_w\!\star\! \piEd\!\!\left(F\right)\!\! -\! \piRd\!\!\left(\!\tilde{\Upsilon}_w\!\star\! F\!\right)\!\right\|_{\infty}
			\!\!\leq \!\tilde{C}\!\left\|w\!-\!\hat{w}\right\|t \!+\! Ch^2t,
		\end{split}
	\end{equation}
	where $\tilde{\Psi}_w$, $\tilde{\Phi}_w$, $\tilde{\Upsilon}_w$, $\piRd$ and $\piEd$
	are defined by (\ref{newPsi_}), (\ref{newPhi_}), (\ref{newUpsilon_}) and (\ref{new_piD}), respectively, $\tilde{C} =24{F_1G_2{F^{-1}_2}C_aC_d}$, $C= (8F_1H_2+2F_2H_1+8G_1G_2){F^{-1}_2}C_a$, $C_d = (p+1)h$ is the diameter constant of filters, $C_a = F_2p^2$ is a constant correlated to the initialization scheme\footnote{In commonly used initialization schemes \cite{fu2024rotation},  $nF_2p^2$ is usually set
as a constant, where $n$ is the number of channels, and here we have $n=1$.}, $\|\cdot\|_{\infty}$ represents the infinity norm.
\end{Thm}

Theorem \ref{Thm_d} demonstrates that the equivariance error for a single convolution layer of the proposed method is primarily influenced by the mesh size $h$ and the estimation error of the learnable transformation parameters $w$. 
Existing rotaion equivariant architectures typically assume ideal rotation symmetry in datasets, which inherently presumes that $w=\hat{w}=[0,1,1]$. As a result, their derived equivariance error bounds only consider terms related to the mesh size $h$ \cite{xie2022fourier, fu2024rotation}. In contrast, we address the more realistic scenario where datasets exhibit non-strict symmetry, thereby necessitating the inclusion of parameter estimation error $\left\|w-\hat{w}\right\|$ in the analysis of equivariance error.
This theoretically demonstrates that proper adaptation to dataset-specific transformations is essential for enhancing the network's equivariance properties.

\subsubsection*{\bf Equivariance Error Across the Entire Network}
We further derive the equivariance error of the entire network and present the following theorem to provide a more comprehensive and precise evaluation of the rotation equivariance error.
\begin{Thm}\label{Thm_entire}
    For an image ${I}$ with size $H\times W\times n_0$, and a $L$-layer equivariant CNN network $\operatorname{CNN}^{eq}_{w}(\cdot)$ constructed by the proposed TL-Conv with transformation-dependent parameters denoted as $w$, whose channel number of the $l^{th}$ layer is $n_l$, and activation function is set as ReLU. If the latent continuous function of the $c^{th}$ channel of $I$ is denoted as $r_c: \mathbb{R}^2 \! \rightarrow \! \mathbb{R}$, and the latent continuous function of any convolution filters in the $l^{th}$ layer is denoted as $\varphi^{l}: \mathbb{R}^2 \! \rightarrow \! \mathbb{R}$, where $l \in \{1, \cdots, L\}$, $c \in \{1, \cdots, n_{0}\}$, for any $x\in \R^2$, the following conditions are satisfied:
	\begin{equation}
		\begin{split}
			& |r_c(x)| \leq F_0, \|\nabla r_c(x)\| \leq G_0, \|\nabla ^2 r_c(x)\| \leq H_0, \\
			& |\varphi^{l}(x)| \leq F_l, \|\nabla \varphi^{l}(x)\| \leq G_l, \|\nabla ^2 \varphi^{l}(x)\| \leq H_l,\\
			&\forall \|x\|\geq\nicefrac{(p+1)h}{2},~ \varphi_{l}(x)=0,\\
		\end{split}
	\end{equation}
	where $p$ is the filter size, $h$ is the mesh size, $\nabla$ and ${\nabla}^2$ denote the operators of gradient and Hessian matrix, respectively. 
	Then for $\forall w,\hat{w}\in \mathbb{R}^3$,  $\tilde{A}\in S$,  $S$ is the rotation subgroup, the following result is satisfied:
	\begin{equation}\label{main_conclusion}
		\begin{split}
			\l|\operatorname{CNN}^{eq}_{w}\!\left[\piRd \right]\! ({I}) \!\!-\! \piEd \!\left[\operatorname{CNN}^{eq}_{w} \right]\!({I})\r|
			\!\leq\! \tilde{C}\!\left\|w\!-\!\hat{w}\right\| \!+ \!Ch^2\!\!,
		\end{split}
	\end{equation}
	where $\piRd$ and $\piEd$ are defined in (\ref{new_piD}), $\tilde{C}\!=\!\!\sum_{l=1}^{L} \!\!\frac{24F_0 C_dG_l\mathcal{F}}{F_l}$ and $C = 8\mathcal{F}\sum_{l=1}^{L} \left(\sum_{m=1}^{l}\frac{H_m F_0}{F_m}   + \sum_{n=1}^{l}\sum_{m=1}^{n-1} \frac{2G_mG_n F_0}{F_mF_n} \right.\\
    \left. +  \sum_{m=1}^{l} \frac{G_m G_0}{F_m}  + \frac{H_0}{4} \right)$,  $\mathcal{F} \!=\prod \nolimits_{l=1}^L n_{l-1} p^2 F_l$ is a constant correlated to the initialization scheme, and  $C_d$ is the diameter constant of filters.
\end{Thm}

Theorem \ref{Thm_entire} provides the bound of the equivariance error with respect to the multi-layer network, which is primarily influenced by the mesh size $h$ and $\left\|w-\hat{w}\right\|$, similar to Theorem \ref{Thm_d}. In practice, the conditions in the above theorem are easy to satisfy, which only need the first and second derivatives of the underlying function of the input images and filters to be bounded. Thus, this demonstrates that the proposed TL-Conv exhibits robust theoretical guarantees regarding its equivariance, with minimal and controllable error in practical applications. 

\section{Experimental Results}\label{experiment}
In this section, we first quantitatively analyze the symmetry properties in image dataset. Our results demonstrate that local features exhibit stronger symmetry under adaptive learnable groups compared to strict rotation groups. 
In subsequent subsections, we evaluate the proposed method across various unpaired image-to-image translation tasks, including both natural and medical images. Besides, we also demonstrate the effectiveness of the proposed TL-Conv on fundamental image restoration tasks, through extensive experiments including rain removal \cite{wang2020model}, image denoising \cite{tian2020deep}, and super-resolution \cite{lim2017enhanced}.

\subsection{Symmetry Analysis of Image Datasets} 
\begin{figure}
    \centering
    \includegraphics[width=\linewidth]{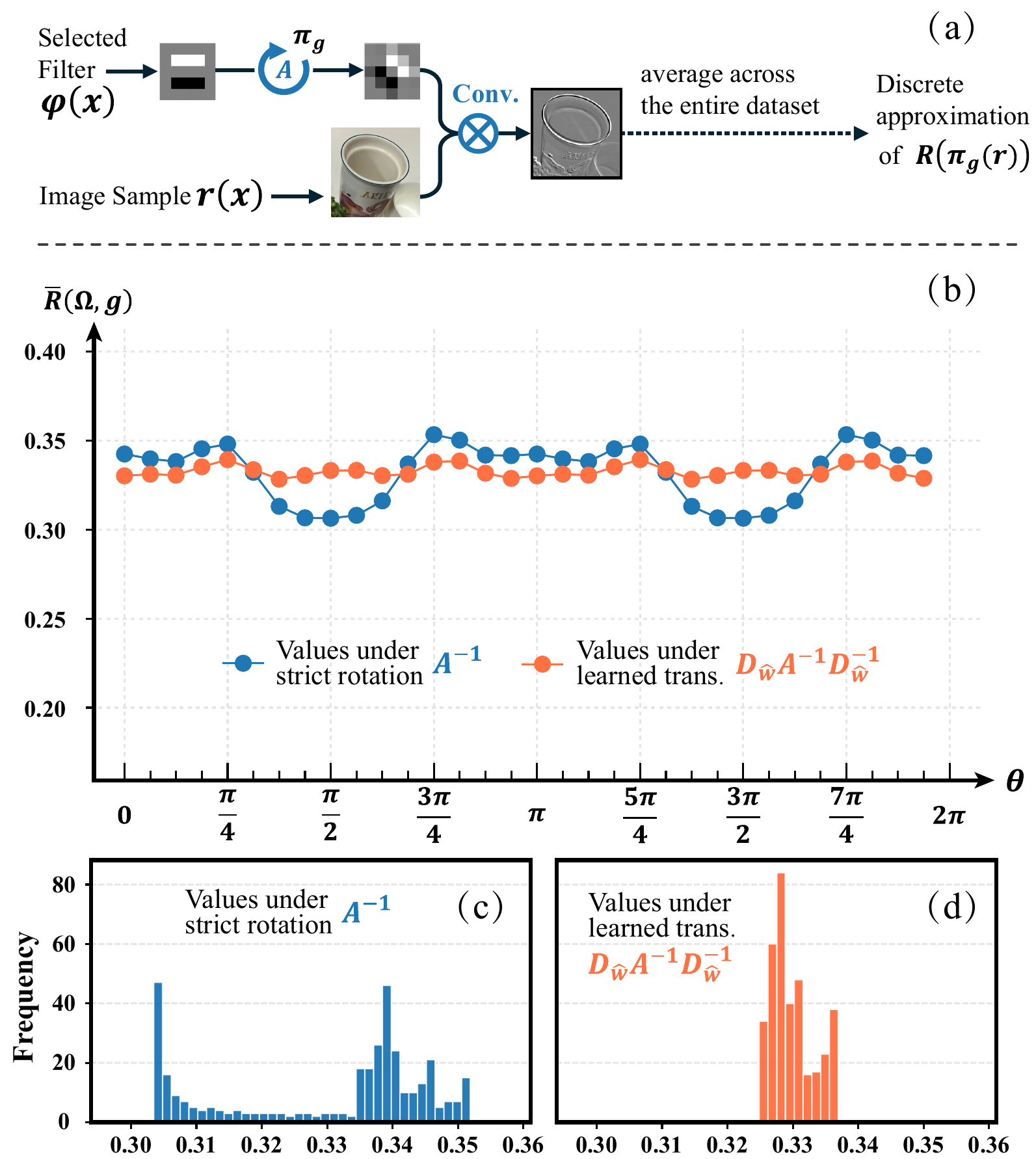}
    \caption{Illustration of data symmetry. (a) The process of feature extraction from images using classical differential kernels. (b) The mean local feature values of testing natural images at different rotation angles. (c)-(d) Illustration of histograms under strict rotation and learned transformations.}
    \label{TV}\vspace{-2mm}
\end{figure}

In Sec. IV-A, we have formally defined the symmetry of local image features, as presented in (\ref{symmetry}). Here, we quantitatively measure the symmetry of the dataset by analyzing the statistical stability of image local features with respect to transformations.
Specifically, for a dataset $\Omega$, we say it is symmetric with respect to a transformation group $G$, if the mean value of feature extraction result is stable for arbitrary $ g\in G$, i.e., $\operatorname{Var}_{g\in G}\!\left(\bar{R}\left(\Omega,g\right)\right)\approx 0$, where
\begin{equation}\label{R_Omega}
    \bar{R}\left(\Omega,g\right) = \frac{1}{|\Omega|}\sum_{r\in\Omega}R(\pi_g (r)).
\end{equation}
Here, we exploit convolution based extractor as an example, which is one of the most typical local feature extractors \cite{hertel2015deep}. Specifically, we have
\begin{equation}\label{R_compute}
    R(r) = \int_{\R^2}\int_{\R^2}\varphi(x)r(y-x)dxdy,
\end{equation}
where $\varphi(x)$ is a filter for feature extraction. In this section, we choose the classical differential kernel for specific calculations. Besides, we can effectively calculate the discrete form of \eqref{R_compute} with the process shown in Fig. \ref{TV}(a)\footnote{It can be proved that $R\left(\pi_g[r]\right) \!\!=\!\! \int_{\R^2}\int_{\R^2}\varphi(g\cdot x)r(y-x)dxdy, \forall g\in S_w$, please refer to the supplementary material for proofs.}.

Based on the above definitions and deductions, we can respectively set $G$ in \eqref{R_Omega} as the strict rotation group $S$ and the proposed learnable group $S_w$, and compare the symmetry of the dataset under these two transformation groups. Particularly, we can optimize the variable $w$ in $\operatorname{Var}_{g\in S_w}\!\left(\bar{R}\left(\Omega,g\right)\right)$ to obtain $\hat{w}$ that yields the optimal symmetry.
Please refer to the supplementary materials for more implementation details.

We employ the widely adopted DIV2K dataset \cite{agustsson2017ntire} as example to conduct the dataset symmetry exploration. As shown in Fig. \ref{TV}(b), it is evident that by introducing learnable transformations (orange part) the distribution of average feature values becomes more stable, as compared to the strict rotation case (blue part). 
Besides, by comparing the histograms in  Fig. \ref{TV}(c) and (d), we can easily observe that the variance of the local feature function under the proposed transformation group is smaller.
These results clearly demonstrate that learnable transformations effectively enhance the intrinsic symmetry of the image data.

\begin{table}[t]
  \caption{Quantiative results on summer2winter\_yosemite and old2young datasets.}
  \label{unpair1}
  \centering \setlength{\tabcolsep}{5.5pt}
  \begin{tabular}{lccc}
    \toprule
       \multirow{3}{*}{Method}  &  {winter$\rightarrow$summer} 
        &  {old$\rightarrow$young}
        &  \multirow{3}{*}{Param. (M)}\\
        \cmidrule(r){2-3}
     & FID $\downarrow$ & FID $\downarrow$ & \\
    \midrule
    NEGCUT \cite{wang2021instance} & 75.8  & 45.8  & -  \\
    MoNCE \cite{zhan2022modulated} & 78.2 & 42.8  & - \\
    QS-Attn \cite{hu2022qs} & 77.2  & 45.2  & - \\
    SRC \cite{jung2022exploring} & 71.6  & 42.7  & - \\
    \midrule
    CycleGAN \cite{zhu2017unpaired} & 72.437 & 41.774 & 11.378 \\
    EQ-CycleGAN & 72.318  & 42.691  & 2.845 \\
    TLEQ-CycleGAN & \textbf{71.504}  & \textbf{41.182}  & 2.845 \\
    \midrule
    CUT \cite{park2020contrastive} & 77.059 & 40.891 & 11.378  \\
    EQ-CUT & 76.460  & 38.747  & 3.541 \\
    TLEQ-CUT & \textbf{73.976}  & \textbf{37.596}  & 3.541 \\
    \midrule
    santa \cite{xie2023unpaired} & 71.339 & 41.900 & 11.428 \\
    EQ-santa & 69.911  & 37.839  & 3.553 \\
    TLEQ-santa & \textbf{69.036}  & \textbf{37.568}  & 3.554 \\
    \bottomrule
  \end{tabular}
\end{table}
\begin{table}
  \caption{Quantiative results on BDD100K dataset.}
  \label{unpair2}
  \centering \setlength{\tabcolsep}{19pt}
  \begin{tabular}{lcc}
    \toprule
      \multirow{3}{*}{Method}  &  {day$\rightarrow$night} 
        &  \multirow{3}{*}{Param. (M)}\\
        \cmidrule(r){2-2}
      & FID $\downarrow$ \\
    \midrule
    CycleGAN \cite{zhu2017unpaired} & 43.507 & 11.378 \\
    EQ-CycleGAN & 38.715  & 2.845 \\
    TLEQ-CycleGAN & \textbf{36.921} & 2.845 \\
    \midrule
    CUT \cite{park2020contrastive} & 35.015 & 11.378  \\
    EQ-CUT & 34.335  & 3.541 \\
    TLEQ-CUT & \textbf{33.599}  & 3.541 \\
    \midrule
    santa \cite{xie2023unpaired} & 35.084  & 11.428 \\
    EQ-santa & 33.569  & 3.553 \\
    TLEQ-santa & \textbf{33.329} & 3.554 \\
    \bottomrule
  \end{tabular}
\end{table}

\begin{figure*}[t]
\vspace{0.5mm}
\hspace{-0mm}\includegraphics[width=1.0\linewidth]{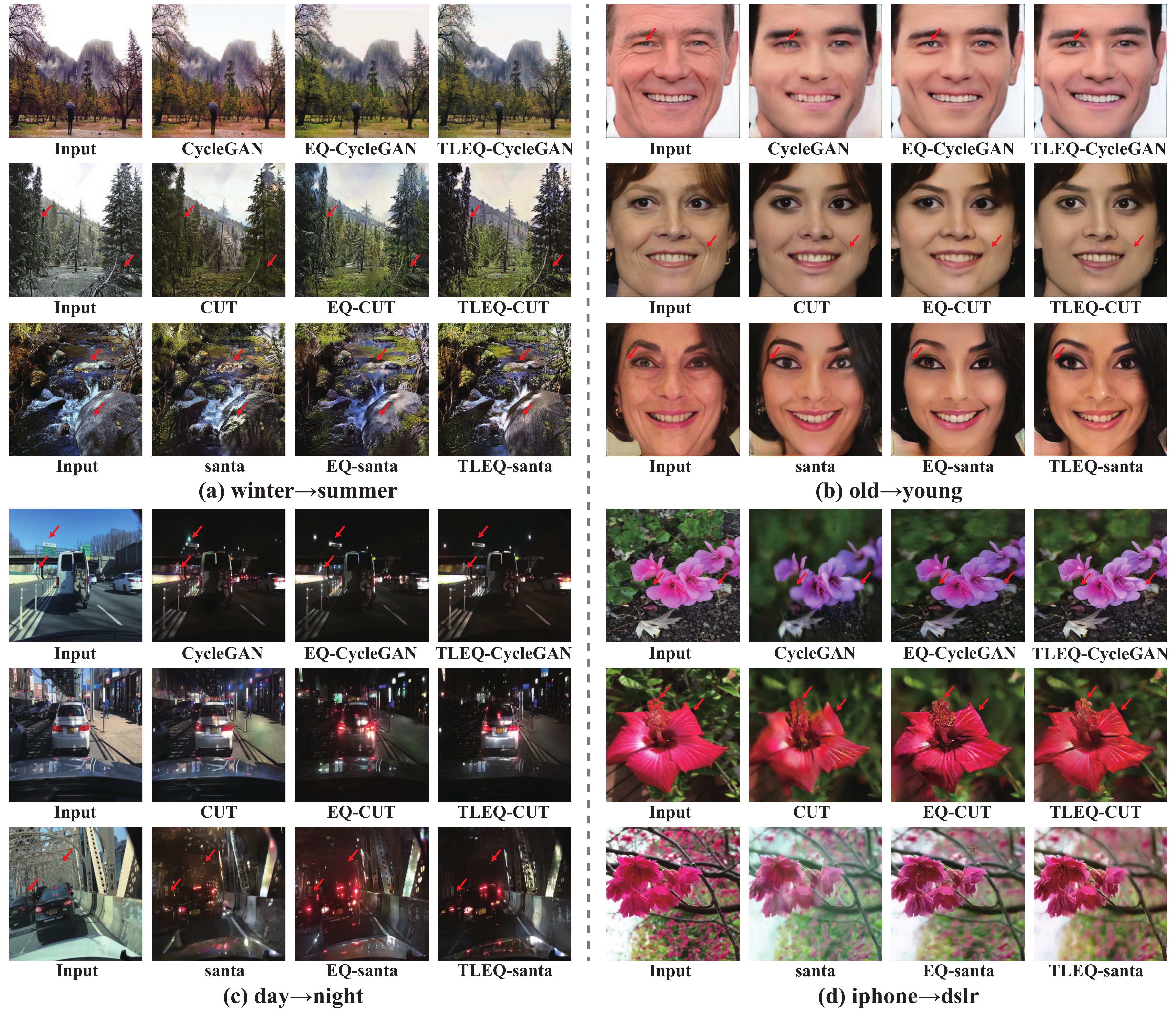}
\vspace{-5mm}
  \caption{The generated samples on four tasks: winter$\rightarrow$summer, old$\rightarrow$young, day$\rightarrow$night and iphone$\rightarrow$dslr. We show the results using CycleGAN, CUT, and santa, along with their equivariant versions (EQ- and TLEQ-).
  Traditional CNN-based methods exhibit several limitations, including insufficient style transfer (e.g., CycleGAN for winter$\rightarrow$summer, CUT for day$\rightarrow$night) and color distortion (santa for iPhone$\rightarrow$dslr). Additionally, CNNs struggle to preserve intrinsic image structures, whereas equivariant networks are better at preserving critical domain-invariant features (see representative examples annotated by red arrows in the figures).}
\label{unpair}
\vspace{-1.5mm}
\end{figure*}

\subsection{Unpaired Image-to-Image Translation}
\subsubsection*{\bf Datasets and Evaluation Metrics}
In this section, we mainly refer to the previous works \cite{zhu2017unpaired, xie2023unpaired, jiang2020tsit}, conducting experiments on the following publicly available datasets: summer2winter\_yosemite \cite{zhu2017unpaired}, old2young \cite{xie2023unpaired}, BDD100K \cite{jiang2020tsit, yu2018bdd100k}, and iphone2dslr\_flower \cite{zhu2017unpaired}.
All images are resized to a size of $256 \times 256$, with training and testing conducted on an NVIDIA 4090 GPU. 
For unpaired problems, we use the widely adopted Frechet Inception Distance (FID) \cite{heusel2017gans} as the quantitative evaluation metric, which is primarily used to measure the similarity between generated images and real images.

\subsubsection*{\bf Network Architecture Settings}
We exploit three well-known image-to-image translation networks: CycleGAN \cite{zhu2017unpaired}, CUT \cite{park2020contrastive}, and santa \cite{xie2023unpaired} as baselines for our experiments. To validate the effectiveness of the proposed methods, we construct corresponding strict rotation equivariant networks (method in Sec. \ref{Rot I2I}) and transformation learnable equivariant networks (proposed in Sec. \ref{TLEConv}) based on the selected CNN architectures in a plug-and-play way\footnote{In practical implementation, referring to B-Conv \cite{xie2025rotation}, we adopt the bicubic functions as the basis functions for filter parametrization to construct the equivariant convolution operators, thereby realizing the equivariant I2I framework.}. These two types of equivariant networks are denoted as EQ- and TLEQ-, respectively. In detail, the rotation group is designed as p4 group ($\nicefrac{k\pi}{2} \mbox{ rotaions, } k = 1,2,3,4$) in all experiments. Therefore, when using the same number of channels, the equivariant convolutions only require $\nicefrac{1}{4}$ filters and thus tend to have fewer network parameters (see Table \ref{unpair1} and Table \ref{unpair2}). The training settings and loss functions are set the same as the original CNN-based methods for fair comparison. For more details, please refer to our released codes.

\begin{figure*}[t]
\vspace{0.5mm}
\hspace{-0mm}\includegraphics[width=1.0\linewidth]{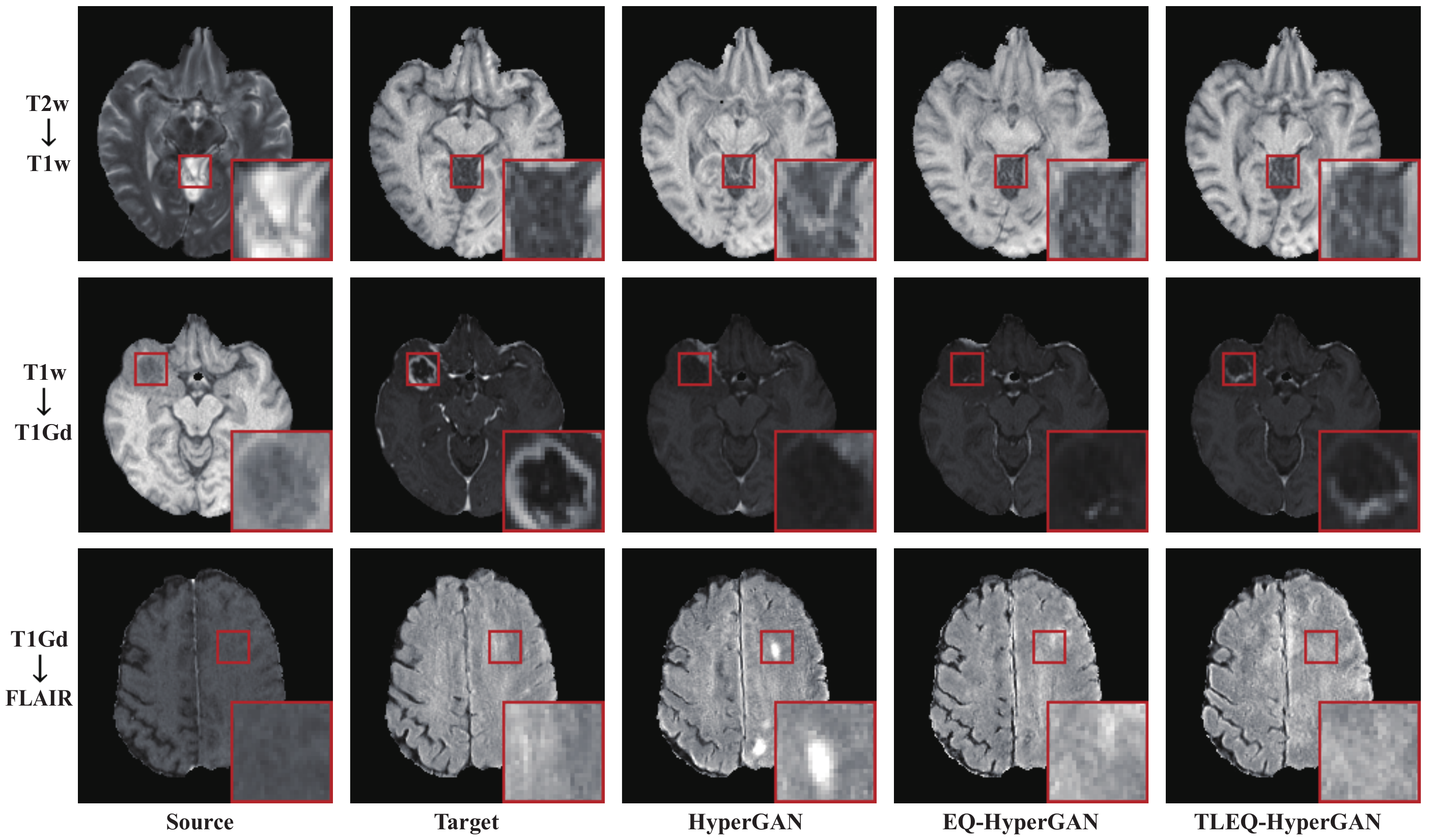}
\vspace{-5mm}
  \caption{Visual comparison of translation results for three randomly selected pairs from the four MRI modalities in the BraTS 2019 dataset. For easy observation, the demarcated regions are zoomed in three times.} 
\label{MRI}
\vspace{-1.5mm}
\end{figure*}

\subsubsection*{\bf Comparison with Typical Methods}
The quantitative results are shown in Table \ref{unpair1} and Table \ref{unpair2}\footnote{For iphone$\rightarrow$dslr task, we follow Zhu et al. \cite{zhu2017unpaired}, which only presents a selection of the most successful samples from this dataset, without calculating the metrics.}. It should be noted that EQ-CycleGAN denotes the strict rotation equivariant variant of CycleGAN, while TLEQ-CycleGAN represents its transformation learnable equivariant counterpart. The naming convention for other networks follows a similar pattern. 
It is evident that the strict rotation equivariant networks outperform standard CNNs across most tasks, generating results that are closer to the true image distribution. 
When further introducing adaptive learning, the proposed transformation learnable equivariant networks lead to additional performance improvements, highlighting the advantages of adaptive learning.
Among these, TLEQ-santa achieves the best results, with an FID of $69.036$ in winter$\rightarrow$summer task and $37.568$ in old$\rightarrow$young. For the day$\rightarrow$night task, TLEQ-santa achieves the lowest FID of $33.329$, and TLEQ-CycleGAN shows a substantial reduction of $6.586$ in FID compared to the standard CycleGAN. These results provide compelling evidence that the proposed method significantly enhances network performance.

Figure \ref{unpair} visually shows the results of three baselines and their corresponding equivariant versions. 
It is evident that CNNs may struggle to produce adequate style transfer (CycleGAN in Fig. \ref{unpair}(a), CUT in Fig. \ref{unpair}(b) and Fig. \ref{unpair}(c)), and potentially result in issues such as unrealistic artifacts (CycleGAN and santa in Fig. \ref{unpair}(c)) and color distortion (santa in Fig. \ref{unpair}(d)). 
In addition, CNNs exhibit fundamental constraints in preserving image structures. For instance, they may produce unintended grid-like artifacts during season translations (CUT in Fig. \ref{unpair}(a)), damage key details such as the eyes and eyebrows during facial rejuvenation (CycleGAN and santa in Fig. \ref{unpair}(b)), or distort the edges of objects when blurring floral backgrounds (CycleGAN and CUT in Fig. \ref{unpair}(d)). In contrast, proposed methods are better at preserving domain-invariant features. Representative examples are highlighted by red arrows in Fig. \ref{unpair}.

\subsection{Multi-contrast MR Image Translation}
Magnetic resonance imaging (MRI) is extensively used in various practical applications. These images typically consist of multiple modalities that assist clinicians in diagnosing and analyzing different tissue regions. The primary MR contrasts include T1-weighted (T1w), T1 with gadolinium enhancement (T1Gd), T2-weighted (T2w), and fluid-attenuated inversion recovery (FLAIR). Due to the high cost of data acquisition, cross-contrast MRI synthesis has become an important area of research. Many studies have focused on generating missing modalities from available ones, with most approaches being one-to-one cross-contrast image synthesis \cite{dar2019image, huang2020mcmt, li2023multi}. 

\begin{table}[t]
  \caption{Accuracies of different methods for arbitrary cross-contrast MRI translation on BraTS 2019 dataset.}
  \label{table_MRI}
  \centering \setlength{\tabcolsep}{7pt}
  \begin{tabular}{lcccc}
    \toprule
    Method  & MAE $\downarrow$ & PSNR $\uparrow$  & SSIM $\uparrow$  & Param.(M) \\
    \midrule
    StarGAN \cite{choi2018stargan} & 0.0083 & 30.75 & 0.905 & 50.72 \\
    DGGAN \cite{tang2018dual} & 0.0081  & 30.81  & 0.908 & 58.74  \\
    SUGAN \cite{sohail2019unpaired} & 0.0087  & 30.37  & 0.909  & 82.63  \\
    ComboGAN \cite{anoosheh2018combogan} & 0.0081  & 30.55  & 0.913  & 71.02  \\
    \midrule
    HyperGAN \cite{yang2021unified} & 0.0067 & 32.30 & 0.9320 & 12.90 \\
    EQ-HyperGAN  & 0.0064  & 32.54  & 0.9362  & 3.08  \\
    TLEQ-HyperGAN & \textbf{0.0062}  & \textbf{32.77}  & \textbf{0.9365}  & 6.09 \\
    \bottomrule
  \end{tabular}
\end{table}

Yang et al. \cite{yang2021unified} introduced a unified HyperGAN model that utilizes a shared encoder and decoder, with a modulator based on one-hot encoding and a multi-layer perceptron (MLP) architecture to control the modality type of the network output. This model efficiently achieves translations between any two MRI contrasts. 
We employ this architecture as our baseline model, subsequently developing the EQ-HyperGAN and TLEQ-HyperGAN variants to validate the effectiveness of the proposed equivariant I2I framework in the medical imaging domain. 
To ensure a fair comparison, the experiments are conducted using the publicly available BraTS 2019 dataset \footnote{https://www.med.upenn.edu/cbica/brats2019.html}, which is employed in Yang et al \cite{yang2021unified}. The data preprocessing operation and all other experimental settings are kept consistent with those used in their study.

We use 2D axial slices of original MR images for training and compute mean absolute error (MAE), peak signal-to-noise ratio (PSNR), and structural similarity (SSIM) between 3D volumes of ground truth and image translation results (all metrics are averaged over the test set and all $12$ translation tasks). 
As shown in Table \ref{table_MRI}, EQ-HyperGAN outperforms CNN-based HyperGAN, with the transformation learnable TLEQ-HyperGAN establishing new state-of-the-art results across all evaluation metrics.
Specifically, TLEQ-HyperGAN achieves a PSNR of $32.77$ dB, representing $0.47$ dB improvement over conventional CNN methods and a $0.23$ dB gain over the strict rotation equivariant approach.
Fig. \ref{MRI} presents a selection of translation tasks as examples for visualization. The results show that the proposed method generates more realistic contrast and better preserves local details, which is crucial for medical diagnosis.


\begin{table}
\centering
\caption{The rain removal results of different competing methods on Rain100L dataset.}

\setlength{\tabcolsep}{24pt}
\begin{tabular}{ccc}
    \toprule
  \multirow{2}{*}{Method} & \multicolumn{2}{c}{Rain100L \cite{yang2019joint}} \\
  \cmidrule(r){2-3}
  & PSNR & SSIM \\
  \midrule
  Input & 26.90 & 0.8384 \\
  DSC\cite{luo2015removing} & 27.34 & 0.8494 \\
  GMM\cite{li2016rain} & 29.05 & 0.8717 \\
  JCAS\cite{gu2017joint}  & 28.54 & 0.8524 \\
  Clear\cite{fu2017clearing} &30.24 & 0.9344 \\
  DDN\cite{fu2017removing} & 32.38 & 0.9258 \\
  RESCAN\cite{li2018recurrent} & 38.52& 0.9812 \\
  PReNet\cite{ren2019progressive} & 37.45 & 0.9790 \\
  SPANet\cite{wang2019spatial} & 35.33 & 0.9694 \\
  JORDER\_E\cite{yang2019joint} & 38.59 & 0.9834 \\
  SIRR\cite{wei2019semi} & 32.37 & 0.9258 \\
  IDT\cite{xiao2022image} & 35.42 & 0.9674 \\
  \midrule
  RCDNet\cite{wang2020model} & 40.00 & 0.9860 \\
  EQ-RCDNet & 40.25 & 0.9863 \\
  TLEQ-RCDNet & \textbf{40.36} & \textbf{0.9869} \\
  \bottomrule
\end{tabular}\vspace{-2mm}
\label{table_derain}
\end{table}

\subsection{Extended Experiments on Image Restoration}
Image restoration is a crucial task in computer vision, focused on improving the quality of degraded images. 
Common tasks include rain removal, denoising, super-resolution, and so on. All of which are essential in a wide range of applications, such as satellite image processing, medical imaging, and surveillance. In this section, we conduct experiments on image restoration tasks to further validate the potential of the proposed methods. Due to space constraints, we have only provided numerical results, while more visual comparison with previous methods can be found in the supplementary material.

\subsubsection*{\bf Single Image Rain Removal}
It is a challenging task in image processing, aiming at eliminating rain streaks from a rainy image while preserving the underlying scene details. Wang et al. \cite{wang2020model} proposed a deep unfolding model named RCDNet, which achieves state-of-the-art performance. 
Similar to the experiments in the aforementioned sections, we construct EQ-RCDNet and TLEQ-RCDNet for comparative analysis.
All experiments are conducted on Rain100L \cite{yang2019joint}, a widely used benchmark dataset, with the remaining settings consistent with the original RCDNet. 

As shown in Table \ref{table_derain}, RCDNet achieves superior performance compared to other CNN-based methods, significantly improving the PSNR and SSIM scores. We can observe that when transformed into the equivariant network, the quality of the generated images further enhances. In particular, the proposed TLEQ-RCDNet achieves state-of-the-art performance. The PSNR is improved respectively by $0.36$ dB and $0.11$ dB compared to the original RCDNet and EQ-RCDNet. 
This clearly demonstrates that the rotation equivariant network makes more effective usage of prior knowledge about the images, resulting in higher-quality outputs. 

\subsubsection*{\bf Image Denoising}
\begin{table*}[t]
    \renewcommand\arraystretch{1.0}
      \centering
            \caption{The denoising results of different competing methods on synthesized testing datasets.} 
               \setlength{\tabcolsep}{15pt}
      \begin{tabular}{c c c c c c c c c c }
         \toprule
        \multirow{2}{*}{Method}  & \multicolumn{2}{c}{Urban100 \cite{huang2015single}} & \multicolumn{2}{c}{BSD100 \cite{martin2001database}} & \multicolumn{2}{c}{Set14 \cite{zeyde2012single}} & \multicolumn{2}{c}{Set5 \cite{bevilacqua2012low}}   \\
        \cmidrule(r){2-3}\cmidrule(r){4-5} \cmidrule(r){6-7}\cmidrule(r){8-9}
         & PSNR & SSIM & PSNR & SSIM & PSNR & SSIM & PSNR & SSIM  \\
         \midrule
        CNN & 30.63 & 0.8864 & 29.56 & 0.8167 & 29.88 & 0.8106 & 31.86 & 0.8803 \\
        SCUNet \cite{zhang2022practical} & 30.50 & 0.8887 & 29.55 & 0.8176 & 29.96 & \textbf{0.8159} & 31.79 & 0.8805\\
        G-CNN \cite{cohen2016group}  & 30.75 & 0.8914 & 29.60  & 0.8199 & 29.89 & 0.8111 & 31.91 & 0.8825 \\
        E2-CNN \cite{weiler2018learning}  & 30.66 & 0.8891 & 29.58 & 0.8177 & 29.79 & 0.8086 & 31.90 & 0.8817 \\
        PDO-eConv \cite{shen2020pdo} & 29.55 & 0.8628 & 29.19 & 0.7998 & 29.50 & 0.8030 & 31.47 & 0.8703 \\
        F-Conv \cite{xie2022fourier}  & 31.05 & \textbf{0.8962} & 29.66 & 0.8211 & 30.02 & 0.8134 & 31.98 & 0.8836 \\
        B-Conv \cite{xie2025rotation}  & 30.97 & 0.8924 & 29.61 & 0.8167 & 29.99 & 0.8113 & 31.94 & 0.8806 \\
        TL-Conv  & \textbf{31.06} & 0.8956 & \textbf{29.67} & \textbf{0.8211} & \textbf{30.05} & 0.8145 & \textbf{32.02} & \textbf{0.8838} \\
       \bottomrule

      \end{tabular}
      \label{table_denoising}
\end{table*}
It aims to remove noise from a corrupted image while preserving its important features, such as edges and textures. 
Following the work of Fu et al. \cite{fu2024rotation}, we compare with several methods, including standard CNN, Swin-Conv UNet (SCUNet) \cite{zhang2022practical} and five state-of-the-art equivariant convolution methods: G-CNN \cite{cohen2016group}, E2-CNN \cite{weiler2018learning}, PDO-eConv \cite{shen2020pdo}, F-Conv \cite{xie2022fourier}, and B-Conv \cite{xie2025rotation}, along with the proposed TL-Conv. All competing methods use the same backbone, i.e., a ResNet \cite{he2016deep} with 16 residual blocks, each containing 256 channels.
For G-CNN, we set the equivariant number to $4$ as it is designed for this configuration, while for the other equivariant convolutions, the equivariant number is $8$. All experimental settings are based on the methodology outlined in the work of Fu et al. \cite{fu2024rotation}. The training dataset used is DIV2K \cite{agustsson2017ntire}, with the test datasets including Set5 \cite{bevilacqua2012low}, Set14 \cite{zeyde2012single}, BSD100 \cite{martin2001database}, and Urban100 \cite{huang2015single}.

As shown in Table \ref{table_denoising}, most of the equivariant methods demonstrate a performance advantage when compared to both the standard convolutional network (CNN) and the Swin transformer-based network (SCUNet). 
When compared to other equivariant methods, the proposed TL-Conv achieves consistent improvements. 

\subsubsection*{\bf Image Super-Resolution}
\begin{table*}[t]
    \renewcommand\arraystretch{1.0}
      \centering
            \caption{The SR results of different competing methods on 4 exploited image datasets.} 
               \setlength{\tabcolsep}{15pt}
      \begin{tabular}{c c c c c c c c c c }
         \toprule
        \multirow{2}{*}{Method}  & \multicolumn{2}{c}{Urban100 \cite{huang2015single}} & \multicolumn{2}{c}{BSD100 \cite{martin2001database}} & \multicolumn{2}{c}{Set14 \cite{zeyde2012single}} & \multicolumn{2}{c}{Set5 \cite{bevilacqua2012low}}   \\
        \cmidrule(r){2-3}\cmidrule(r){4-5} \cmidrule(r){6-7}\cmidrule(r){8-9}
         & PSNR & SSIM & PSNR & SSIM & PSNR & SSIM & PSNR & SSIM  \\
         \midrule
        CNN & 32.17 & 0.9297 & 32.20 & 0.9029 & 33.62 & 0.9198 & 38.06 & 0.9621 \\
        Har-Net \cite{zhang2022practical} & 31.99 & 0.9281 & 32.16 & 0.9025 & 33.53 & 0.9192 & 37.92 & 0.9619\\
        G-CNN \cite{cohen2016group}  & 32.27 & 0.9306 & 32.21  & 0.9027 & 33.61 & 0.9197 & 38.04 & 0.9621 \\
        E2-CNN \cite{weiler2018learning}  & 31.99 & 0.9283 & 32.15 & 0.9024 & 33.51 & 0.9191 & 37.91 & 0.9618 \\
        PDO-eConv \cite{shen2020pdo} & 30.91 & 0.9159 & 31.87 & 0.8985 & 33.17 & 0.9161 & 37.63 & 0.9606 \\
        F-Conv \cite{xie2022fourier}  & 32.36 & 0.9313 & 32.24 & 0.9034 & 33.70 & 0.9205 & 38.06 & 0.9622 \\
        B-Conv \cite{xie2025rotation}  & 32.49 & 0.9330 & 32.26 & 0.9038 & 33.73 & 0.9208 & 38.12 & 0.9625 \\
        TL-Conv  & \textbf{32.55} & \textbf{0.9337} & \textbf{32.28} & \textbf{0.9039} & \textbf{33.80} & \textbf{0.9217} & \textbf{38.13} & \textbf{0.9625} \\
       \bottomrule

      \end{tabular}
      \label{table_SR}
\end{table*}
Single image super-resolution (SR) focuses on reconstructing a high-resolution image from a single low-resolution input.
Our experiments are based on $2$ scale SR case, following the setup of Xie et al. \cite{xie2022fourier}, and use EDSR \cite{lim2017enhanced} as a baseline which has shown excellent performance in super-resolution tasks. The number of residual blocks is set to $16$, with each residual block having $256$ channels. We compare the following equivariant methods: G-CNN \cite{cohen2016group} (using the p4 group for the equivariant convolutions), Har-Net \cite{worrall2017harmonic}, E2-CNN \cite{weiler2019general}, PDO-eConv \cite{shen2020pdo}, F-Conv \cite{xie2022fourier}, and B-Conv \cite{xie2025rotation} (all using the p8 group for the equivariant convolutions), along with the proposed TL-Conv. For training, we utilize $800$ images from the DIV2K dataset \cite{agustsson2017ntire}. For testing, we employ four standard benchmark datasets: Set5 \cite{bevilacqua2012low}, Set14 \cite{zeyde2012single}, BSD100 \cite{martin2001database}, and Urban100 \cite{huang2015single}. All other settings are kept consistent with Xie et al. \cite{xie2022fourier}.

Table \ref{table_SR} presents SR results of $8$ different methods across four benchmark datasets, using PSNR and SSIM as evaluation metrics. 
Compared to the SOTA strict rotation equivariant convolutions (F-Conv and B-Conv), the proposed TL-Conv further enhances performance, with the highest improvement observed on the Urban100 dataset, which achieves a PSNR increase of $0.19$ dB over F-Conv. 

\section{Conclusion}\label{conclusion}

In this work, we have explored the transformation symmetry prior in image datasets, and focus on constructing I2I frameworks that preserve this domain-invariant feature. Specifically, we have introduced two key components to advance I2I methods: 1) rotation equivariant I2I framewok constructed with existing EQ-CNNs, which preserves strict rotation symmetry across the entire network flow, and 2) a new proposed transformation learnable equivariant convolution named TL-Conv,  which can adaptively learn more symmetric transformations in the dataset and ensure the corresponding equivariance in the I2I process. 
The comprehensive experiments across diverse I2I tasks, including natural image conversion, face rejuvenation, and medical image translation, validate the effectiveness of our proposed method. Additional evaluations on image restoration further highlight its potential to enhance performance in various image processing applications. 

While the proposed transformation learnable equivariant networks improve the adaptation to diverse datasets, this also incurs increased training time. Future work could optimize this computational overhead through lightweight methods or dynamic sparse training strategies. Besides, although our current framework focuses on CNN-based designs, extending transformer-based learnable equivariant methods presents a promising research direction, as vision transformers have shown significant promise in various image processing tasks.
A more comprehensive and deep exploration of transformation learning is also an important issue. In this study, transformation learning was applied globally across the entire dataset. Introducing hyper-networks to predict instance-specific transformations could enable more precise symmetry adaptation, potentially improving the model’s performance.


\bibliographystyle{unsrt}
\bibliography{egbib}

@inproceedings{isola2017image,
  title={Image-to-image translation with conditional adversarial networks},
  author={Isola, Phillip and Zhu, Jun-Yan and Zhou, Tinghui and Efros, Alexei A},
  booktitle={Proceedings of the IEEE conference on computer vision and pattern recognition},
  pages={1125--1134},
  year={2017}
}

@inproceedings{chen2022vector,
  title={Vector quantized image-to-image translation},
  author={Chen, Yu-Jie and Cheng, Shin-I and Chiu, Wei-Chen and Tseng, Hung-Yu and Lee, Hsin-Ying},
  booktitle={European Conference on Computer Vision},
  pages={440--456},
  year={2022},
  organization={Springer}
}

@article{wang2023quantitative,
  title={Quantitative cerebral blood volume image synthesis from Standard MRI using image-to-image translation for brain tumors},
  author={Wang, Bao and Pan, Yongsheng and Xu, Shangchen and Zhang, Yi and Ming, Yang and Chen, Ligang and Liu, Xuejun and Wang, Chengwei and Liu, Yingchao and Xia, Yong},
  journal={Radiology},
  volume={308},
  number={2},
  pages={e222471},
  year={2023},
  publisher={Radiological Society of North America}
}

@inproceedings{deng2022stytr2,
  title={Stytr2: Image style transfer with transformers},
  author={Deng, Yingying and Tang, Fan and Dong, Weiming and Ma, Chongyang and Pan, Xingjia and Wang, Lei and Xu, Changsheng},
  booktitle={Proceedings of the IEEE/CVF conference on computer vision and pattern recognition},
  pages={11326--11336},
  year={2022}
}

@article{sun2023umgan,
  title={UMGAN: Underwater image enhancement network for unpaired image-to-image translation},
  author={Sun, Boyang and Mei, Yupeng and Yan, Ni and Chen, Yingyi},
  journal={Journal of Marine Science and Engineering},
  volume={11},
  number={2},
  pages={447},
  year={2023},
  publisher={MDPI}
}

@inproceedings{zhu2017unpaired,
  title={Unpaired image-to-image translation using cycle-consistent adversarial networks},
  author={Zhu, Jun-Yan and Park, Taesung and Isola, Phillip and Efros, Alexei A},
  booktitle={Proceedings of the IEEE international conference on computer vision},
  pages={2223--2232},
  year={2017}
}

@inproceedings{park2020contrastive,
  title={Contrastive learning for unpaired image-to-image translation},
  author={Park, Taesung and Efros, Alexei A and Zhang, Richard and Zhu, Jun-Yan},
  booktitle={Computer Vision--ECCV 2020: 16th European Conference, Glasgow, UK, August 23--28, 2020, Proceedings, Part IX 16},
  pages={319--345},
  year={2020},
  organization={Springer}
}

@inproceedings{xie2023unpaired,
  title={Unpaired image-to-image translation with shortest path regularization},
  author={Xie, Shaoan and Xu, Yanwu and Gong, Mingming and Zhang, Kun},
  booktitle={Proceedings of the IEEE/CVF Conference on Computer Vision and Pattern Recognition},
  pages={10177--10187},
  year={2023}
}

@inproceedings{huang2023quantart,
  title={Quantart: Quantizing image style transfer towards high visual fidelity},
  author={Huang, Siyu and An, Jie and Wei, Donglai and Luo, Jiebo and Pfister, Hanspeter},
  booktitle={Proceedings of the IEEE/CVF Conference on Computer Vision and Pattern Recognition},
  pages={5947--5956},
  year={2023}
}

@inproceedings{zheng2024puff,
  title={Puff-Net: Efficient Style Transfer with Pure Content and Style Feature Fusion Network},
  author={Zheng, Sizhe and Gao, Pan and Zhou, Peng and Qin, Jie},
  booktitle={Proceedings of the IEEE/CVF Conference on Computer Vision and Pattern Recognition},
  pages={8059--8068},
  year={2024}
}

@inproceedings{weiler2018learning,
  title={Learning steerable filters for rotation equivariant cnns},
  author={Weiler, Maurice and Hamprecht, Fred A and Storath, Martin},
  booktitle={Proceedings of the IEEE Conference on Computer Vision and Pattern Recognition},
  pages={849--858},
  year={2018}
}

@article{weiler2019general,
  title={General e (2)-equivariant steerable cnns},
  author={Weiler, Maurice and Cesa, Gabriele},
  journal={Advances in neural information processing systems},
  volume={32},
  year={2019}
}

@inproceedings{cohen2016group,
  title={Group equivariant convolutional networks},
  author={Cohen, Taco and Welling, Max},
  booktitle={International conference on machine learning},
  pages={2990--2999},
  year={2016},
  organization={PMLR}
}

@article{xie2022fourier,
  title={Fourier series expansion based filter parametrization for equivariant convolutions},
  author={Xie, Qi and Zhao, Qian and Xu, Zongben and Meng, Deyu},
  journal={IEEE Transactions on Pattern Analysis and Machine Intelligence},
  volume={45},
  number={4},
  pages={4537--4551},
  year={2022},
  publisher={IEEE}
}

@article{fu2024rotation,
  title={Rotation Equivariant Proximal Operator for Deep Unfolding Methods in Image Restoration},
  author={Fu, Jiahong and Xie, Qi and Meng, Deyu and Xu, Zongben},
  journal={IEEE Transactions on Pattern Analysis and Machine Intelligence},
  year={2024},
  publisher={IEEE}
}

@inproceedings{shen2020pdo,
  title={Pdo-econvs: Partial differential operator based equivariant convolutions},
  author={Shen, Zhengyang and He, Lingshen and Lin, Zhouchen and Ma, Jinwen},
  booktitle={International Conference on Machine Learning},
  pages={8697--8706},
  year={2020},
  organization={PMLR}
}

@article{celledoni2021equivariant,
  title={Equivariant neural networks for inverse problems},
  author={Celledoni, Elena and Ehrhardt, Matthias J and Etmann, Christian and Owren, Brynjulf and Sch{\"o}nlieb, Carola-Bibiane and Sherry, Ferdia},
  journal={Inverse Problems},
  volume={37},
  number={8},
  pages={085006},
  year={2021},
  publisher={IOP Publishing}
}

@article{pang2021image,
  title={Image-to-image translation: Methods and applications},
  author={Pang, Yingxue and Lin, Jianxin and Qin, Tao and Chen, Zhibo},
  journal={IEEE Transactions on Multimedia},
  volume={24},
  pages={3859--3881},
  year={2021},
  publisher={IEEE}
}

@inproceedings{wang2018discriminative,
  title={Discriminative region proposal adversarial networks for high-quality image-to-image translation},
  author={Wang, Chao and Zheng, Haiyong and Yu, Zhibin and Zheng, Ziqiang and Gu, Zhaorui and Zheng, Bing},
  booktitle={Proceedings of the European conference on computer vision (ECCV)},
  pages={770--785},
  year={2018}
}

@inproceedings{zhang2020cross,
  title={Cross-domain correspondence learning for exemplar-based image translation},
  author={Zhang, Pan and Zhang, Bo and Chen, Dong and Yuan, Lu and Wen, Fang},
  booktitle={Proceedings of the IEEE/CVF conference on computer vision and pattern recognition},
  pages={5143--5153},
  year={2020}
}

@inproceedings{zhou2021cocosnet,
  title={Cocosnet v2: Full-resolution correspondence learning for image translation},
  author={Zhou, Xingran and Zhang, Bo and Zhang, Ting and Zhang, Pan and Bao, Jianmin and Chen, Dong and Zhang, Zhongfei and Wen, Fang},
  booktitle={Proceedings of the IEEE/CVF conference on computer vision and pattern recognition},
  pages={11465--11475},
  year={2021}
}

@article{huang2018introduction,
  title={An introduction to image synthesis with generative adversarial nets},
  author={Huang, He and Yu, Philip S and Wang, Changhu},
  journal={arXiv preprint arXiv:1803.04469},
  year={2018}
}

@article{jing2019neural,
  title={Neural style transfer: A review},
  author={Jing, Yongcheng and Yang, Yezhou and Feng, Zunlei and Ye, Jingwen and Yu, Yizhou and Song, Mingli},
  journal={IEEE transactions on visualization and computer graphics},
  volume={26},
  number={11},
  pages={3365--3385},
  year={2019},
  publisher={IEEE}
}

@article{kaji2019overview,
  title={Overview of image-to-image translation by use of deep neural networks: denoising, super-resolution, modality conversion, and reconstruction in medical imaging},
  author={Kaji, Shizuo and Kida, Satoshi},
  journal={Radiological physics and technology},
  volume={12},
  number={3},
  pages={235--248},
  year={2019},
  publisher={Springer}
}

@article{tian2020deep,
  title={Deep learning on image denoising: An overview},
  author={Tian, Chunwei and Fei, Lunke and Zheng, Wenxian and Xu, Yong and Zuo, Wangmeng and Lin, Chia-Wen},
  journal={Neural Networks},
  volume={131},
  pages={251--275},
  year={2020},
  publisher={Elsevier}
}

@inproceedings{zhao2020uctgan,
  title={Uctgan: Diverse image inpainting based on unsupervised cross-space translation},
  author={Zhao, Lei and Mo, Qihang and Lin, Sihuan and Wang, Zhizhong and Zuo, Zhiwen and Chen, Haibo and Xing, Wei and Lu, Dongming},
  booktitle={Proceedings of the IEEE/CVF conference on computer vision and pattern recognition},
  pages={5741--5750},
  year={2020}
}

@inproceedings{chen2021dualast,
  title={Dualast: Dual style-learning networks for artistic style transfer},
  author={Chen, Haibo and Zhao, Lei and Wang, Zhizhong and Zhang, Huiming and Zuo, Zhiwen and Li, Ailin and Xing, Wei and Lu, Dongming},
  booktitle={Proceedings of the IEEE/CVF conference on computer vision and pattern recognition},
  pages={872--881},
  year={2021}
}

@article{li2023multi,
  title={Multi-scale transformer network with edge-aware pre-training for cross-modality MR image synthesis},
  author={Li, Yonghao and Zhou, Tao and He, Kelei and Zhou, Yi and Shen, Dinggang},
  journal={IEEE Transactions on Medical Imaging},
  volume={42},
  number={11},
  pages={3395--3407},
  year={2023},
  publisher={IEEE}
}

@article{zhou2022hrinversion,
  title={HRInversion: High-resolution GAN inversion for cross-domain image synthesis},
  author={Zhou, Peng and Xie, Lingxi and Ni, Bingbing and Liu, Lin and Tian, Qi},
  journal={IEEE Transactions on Circuits and Systems for Video Technology},
  volume={33},
  number={5},
  pages={2147--2161},
  year={2022},
  publisher={IEEE}
}

@inproceedings{lira2020ganhopper,
  title={Ganhopper: Multi-hop gan for unsupervised image-to-image translation},
  author={Lira, Wallace and Merz, Johannes and Ritchie, Daniel and Cohen-Or, Daniel and Zhang, Hao},
  booktitle={Computer Vision--ECCV 2020: 16th European Conference, Glasgow, UK, August 23--28, 2020, Proceedings, Part XXVI 16},
  pages={363--379},
  year={2020},
  organization={Springer}
}

@inproceedings{kotovenko2019content,
  title={Content and style disentanglement for artistic style transfer},
  author={Kotovenko, Dmytro and Sanakoyeu, Artsiom and Lang, Sabine and Ommer, Bjorn},
  booktitle={Proceedings of the IEEE/CVF international conference on computer vision},
  pages={4422--4431},
  year={2019}
}

@article{meng2024multi,
  title={Multi-modal Modality-masked Diffusion Network for Brain MRI Synthesis with Random Modality Missing},
  author={Meng, Xiangxi and Sun, Kaicong and Xu, Jun and He, Xuming and Shen, Dinggang},
  journal={IEEE Transactions on Medical Imaging},
  year={2024},
  publisher={IEEE}
}

@article{taigman2016unsupervised,
  title={Unsupervised cross-domain image generation},
  author={Taigman, Yaniv and Polyak, Adam and Wolf, Lior},
  journal={arXiv preprint arXiv:1611.02200},
  year={2016}
}

@inproceedings{chen2020simple,
  title={A simple framework for contrastive learning of visual representations},
  author={Chen, Ting and Kornblith, Simon and Norouzi, Mohammad and Hinton, Geoffrey},
  booktitle={International conference on machine learning},
  pages={1597--1607},
  year={2020},
  organization={PMLR}
}

@inproceedings{he2020momentum,
  title={Momentum contrast for unsupervised visual representation learning},
  author={He, Kaiming and Fan, Haoqi and Wu, Yuxin and Xie, Saining and Girshick, Ross},
  booktitle={Proceedings of the IEEE/CVF conference on computer vision and pattern recognition},
  pages={9729--9738},
  year={2020}
}

@inproceedings{wang2021instance,
  title={Instance-wise hard negative example generation for contrastive learning in unpaired image-to-image translation},
  author={Wang, Weilun and Zhou, Wengang and Bao, Jianmin and Chen, Dong and Li, Houqiang},
  booktitle={Proceedings of the IEEE/CVF international conference on computer vision},
  pages={14020--14029},
  year={2021}
}

@inproceedings{pizzati2021comogan,
  title={CoMoGAN: continuous model-guided image-to-image translation},
  author={Pizzati, Fabio and Cerri, Pietro and De Charette, Raoul},
  booktitle={Proceedings of the IEEE/CVF Conference on Computer Vision and Pattern Recognition},
  pages={14288--14298},
  year={2021}
}

@inproceedings{esser2021taming,
  title={Taming transformers for high-resolution image synthesis},
  author={Esser, Patrick and Rombach, Robin and Ommer, Bjorn},
  booktitle={Proceedings of the IEEE/CVF conference on computer vision and pattern recognition},
  pages={12873--12883},
  year={2021}
}

@inproceedings{li2018unsupervised,
  title={Unsupervised image-to-image translation with stacked cycle-consistent adversarial networks},
  author={Li, Minjun and Huang, Haozhi and Ma, Lin and Liu, Wei and Zhang, Tong and Jiang, Yugang},
  booktitle={Proceedings of the European conference on computer vision (ECCV)},
  pages={184--199},
  year={2018}
}

@article{smidt2021euclidean,
  title={Euclidean symmetry and equivariance in machine learning},
  author={Smidt, Tess E},
  journal={Trends in Chemistry},
  volume={3},
  number={2},
  pages={82--85},
  year={2021},
  publisher={Elsevier}
}

@inproceedings{lenc2015understanding,
  title={Understanding image representations by measuring their equivariance and equivalence},
  author={Lenc, Karel and Vedaldi, Andrea},
  booktitle={Proceedings of the IEEE conference on computer vision and pattern recognition},
  pages={991--999},
  year={2015}
}

@phdthesis{cohen2021equivariant,
  title={Equivariant convolutional networks},
  author={Cohen, Taco and others},
  year={2021},
  school={Taco Cohen}
}

@article{krizhevsky2012imagenet,
  title={Imagenet classification with deep convolutional neural networks},
  author={Krizhevsky, Alex and Sutskever, Ilya and Hinton, Geoffrey E},
  journal={Advances in neural information processing systems},
  volume={25},
  year={2012}
}

@inproceedings{quiroga2020revisiting,
  title={Revisiting data augmentation for rotational invariance in convolutional neural networks},
  author={Quiroga, Facundo and Ronchetti, Franco and Lanzarini, Laura and Bariviera, Aurelio F},
  booktitle={Modelling and Simulation in Management Sciences: Proceedings of the International Conference on Modelling and Simulation in Management Sciences (MS-18)},
  pages={127--141},
  year={2020},
  organization={Springer}
}

@article{hoogeboom2018hexaconv,
  title={Hexaconv},
  author={Hoogeboom, Emiel and Peters, Jorn WT and Cohen, Taco S and Welling, Max},
  journal={arXiv preprint arXiv:1803.02108},
  year={2018}
}

@inproceedings{zhou2017oriented,
  title={Oriented response networks},
  author={Zhou, Yanzhao and Ye, Qixiang and Qiu, Qiang and Jiao, Jianbin},
  booktitle={Proceedings of the IEEE Conference on Computer Vision and Pattern Recognition},
  pages={519--528},
  year={2017}
}

@inproceedings{marcos2017rotation,
  title={Rotation equivariant vector field networks},
  author={Marcos, Diego and Volpi, Michele and Komodakis, Nikos and Tuia, Devis},
  booktitle={Proceedings of the IEEE International Conference on Computer Vision},
  pages={5048--5057},
  year={2017}
}

@inproceedings{worrall2017harmonic,
  title={Harmonic networks: Deep translation and rotation equivariance},
  author={Worrall, Daniel E and Garbin, Stephan J and Turmukhambetov, Daniyar and Brostow, Gabriel J},
  booktitle={Proceedings of the IEEE conference on computer vision and pattern recognition},
  pages={5028--5037},
  year={2017}
}

@inproceedings{shen2021pdo,
  title={PDO-eS2CNNs: Partial differential operator based equivariant spherical CNNs},
  author={Shen, Zhengyang and Shen, Tiancheng and Lin, Zhouchen and Ma, Jinwen},
  booktitle={Proceedings of the AAAI Conference on Artificial Intelligence},
  volume={35},
  number={11},
  pages={9585--9593},
  year={2021}
}

@inproceedings{li2024affine,
  title={Affine Equivariant Networks Based on Differential Invariants},
  author={Li, Yikang and Qiu, Yeqing and Chen, Yuxuan and He, Lingshen and Lin, Zhouchen},
  booktitle={Proceedings of the IEEE/CVF Conference on Computer Vision and Pattern Recognition},
  pages={5546--5556},
  year={2024}
}

@inproceedings{lim2017enhanced,
  title={Enhanced deep residual networks for single image super-resolution},
  author={Lim, Bee and Son, Sanghyun and Kim, Heewon and Nah, Seungjun and Mu Lee, Kyoung},
  booktitle={Proceedings of the IEEE conference on computer vision and pattern recognition workshops},
  pages={136--144},
  year={2017}
}

@inproceedings{wang2020model,
  title={A model-driven deep neural network for single image rain removal},
  author={Wang, Hong and Xie, Qi and Zhao, Qian and Meng, Deyu},
  booktitle={Proceedings of the IEEE/CVF conference on computer vision and pattern recognition},
  pages={3103--3112},
  year={2020}
}

@inproceedings{yang2021unified,
  title={A unified hyper-GAN model for unpaired multi-contrast MR image translation},
  author={Yang, Heran and Sun, Jian and Yang, Liwei and Xu, Zongben},
  booktitle={Medical Image Computing and Computer Assisted Intervention--MICCAI 2021: 24th International Conference, Strasbourg, France, September 27--October 1, 2021, Proceedings, Part III 24},
  pages={127--137},
  year={2021},
  organization={Springer}
}

@inproceedings{jiang2020tsit,
  title={Tsit: A simple and versatile framework for image-to-image translation},
  author={Jiang, Liming and Zhang, Changxu and Huang, Mingyang and Liu, Chunxiao and Shi, Jianping and Loy, Chen Change},
  booktitle={Computer Vision--ECCV 2020: 16th European Conference, Glasgow, UK, August 23--28, 2020, Proceedings, Part III 16},
  pages={206--222},
  year={2020},
  organization={Springer}
}

@article{yu2018bdd100k,
  title={Bdd100k: A diverse driving video database with scalable annotation tooling},
  author={Yu, Fisher and Xian, Wenqi and Chen, Yingying and Liu, Fangchen and Liao, Mike and Madhavan, Vashisht and Darrell, Trevor and others},
  journal={arXiv preprint arXiv:1805.04687},
  volume={2},
  number={5},
  pages={6},
  year={2018}
}

@article{heusel2017gans,
  title={Gans trained by a two time-scale update rule converge to a local nash equilibrium},
  author={Heusel, Martin and Ramsauer, Hubert and Unterthiner, Thomas and Nessler, Bernhard and Hochreiter, Sepp},
  journal={Advances in neural information processing systems},
  volume={30},
  year={2017}
}

@inproceedings{zhan2022modulated,
  title={Modulated contrast for versatile image synthesis},
  author={Zhan, Fangneng and Zhang, Jiahui and Yu, Yingchen and Wu, Rongliang and Lu, Shijian},
  booktitle={Proceedings of the IEEE/CVF Conference on Computer Vision and Pattern Recognition},
  pages={18280--18290},
  year={2022}
}

@inproceedings{hu2022qs,
  title={Qs-attn: Query-selected attention for contrastive learning in i2i translation},
  author={Hu, Xueqi and Zhou, Xinyue and Huang, Qiusheng and Shi, Zhengyi and Sun, Li and Li, Qingli},
  booktitle={Proceedings of the IEEE/CVF Conference on Computer Vision and Pattern Recognition},
  pages={18291--18300},
  year={2022}
}

@inproceedings{jung2022exploring,
  title={Exploring patch-wise semantic relation for contrastive learning in image-to-image translation tasks},
  author={Jung, Chanyong and Kwon, Gihyun and Ye, Jong Chul},
  booktitle={Proceedings of the IEEE/CVF conference on computer vision and pattern recognition},
  pages={18260--18269},
  year={2022}
}

@article{dar2019image,
  title={Image synthesis in multi-contrast MRI with conditional generative adversarial networks},
  author={Dar, Salman UH and Yurt, Mahmut and Karacan, Levent and Erdem, Aykut and Erdem, Erkut and Cukur, Tolga},
  journal={IEEE transactions on medical imaging},
  volume={38},
  number={10},
  pages={2375--2388},
  year={2019},
  publisher={IEEE}
}

@article{huang2020mcmt,
  title={MCMT-GAN: multi-task coherent modality transferable GAN for 3D brain image synthesis},
  author={Huang, Yawen and Zheng, Feng and Cong, Runmin and Huang, Weilin and Scott, Matthew R and Shao, Ling},
  journal={IEEE Transactions on Image Processing},
  volume={29},
  pages={8187--8198},
  year={2020},
  publisher={IEEE}
}

@inproceedings{choi2018stargan,
  title={Stargan: Unified generative adversarial networks for multi-domain image-to-image translation},
  author={Choi, Yunjey and Choi, Minje and Kim, Munyoung and Ha, Jung-Woo and Kim, Sunghun and Choo, Jaegul},
  booktitle={Proceedings of the IEEE conference on computer vision and pattern recognition},
  pages={8789--8797},
  year={2018}
}

@inproceedings{tang2018dual,
  title={Dual generator generative adversarial networks for multi-domain image-to-image translation},
  author={Tang, Hao and Xu, Dan and Wang, Wei and Yan, Yan and Sebe, Nicu},
  booktitle={Asian Conference on Computer Vision},
  pages={3--21},
  year={2018},
  organization={Springer}
}

@inproceedings{sohail2019unpaired,
  title={Unpaired multi-contrast MR image synthesis using generative adversarial networks},
  author={Sohail, Muhammad and Riaz, Muhammad Naveed and Wu, Jing and Long, Chengnian and Li, Shaoyuan},
  booktitle={International Workshop on Simulation and Synthesis in Medical Imaging},
  pages={22--31},
  year={2019},
  organization={Springer}
}

@inproceedings{anoosheh2018combogan,
  title={Combogan: Unrestrained scalability for image domain translation},
  author={Anoosheh, Asha and Agustsson, Eirikur and Timofte, Radu and Van Gool, Luc},
  booktitle={Proceedings of the IEEE conference on computer vision and pattern recognition workshops},
  pages={783--790},
  year={2018}
}

@article{zhang2022practical,
  title={Practical blind denoising via swin-conv-unet and data synthesis},
  author={Zhang, Kai and Li, Yawei and Liang, Jingyun and Cao, Jiezhang and Zhang, Yulun and Tang, Hao and Timofte, Radu and Van Gool, Luc},
  journal={arXiv e-prints},
  pages={arXiv--2203},
  year={2022}
}

@inproceedings{he2016deep,
  title={Deep residual learning for image recognition},
  author={He, Kaiming and Zhang, Xiangyu and Ren, Shaoqing and Sun, Jian},
  booktitle={Proceedings of the IEEE conference on computer vision and pattern recognition},
  pages={770--778},
  year={2016}
}

@inproceedings{agustsson2017ntire,
  title={Ntire 2017 challenge on single image super-resolution: Dataset and study},
  author={Agustsson, Eirikur and Timofte, Radu},
  booktitle={Proceedings of the IEEE conference on computer vision and pattern recognition workshops},
  pages={126--135},
  year={2017}
}

@article{bevilacqua2012low,
  title={Low-complexity single-image super-resolution based on nonnegative neighbor embedding},
  author={Bevilacqua, Marco and Roumy, Aline and Guillemot, Christine and Alberi-Morel, Marie Line},
  year={2012},
  publisher={BMVA press}
}

@inproceedings{zeyde2012single,
  title={On single image scale-up using sparse-representations},
  author={Zeyde, Roman and Elad, Michael and Protter, Matan},
  booktitle={Curves and Surfaces: 7th International Conference, Avignon, France, June 24-30, 2010, Revised Selected Papers 7},
  pages={711--730},
  year={2012},
  organization={Springer}
}

@inproceedings{martin2001database,
  title={A database of human segmented natural images and its application to evaluating segmentation algorithms and measuring ecological statistics},
  author={Martin, David and Fowlkes, Charless and Tal, Doron and Malik, Jitendra},
  booktitle={Proceedings eighth IEEE international conference on computer vision. ICCV 2001},
  volume={2},
  pages={416--423},
  year={2001},
  organization={IEEE}
}

@inproceedings{huang2015single,
  title={Single image super-resolution from transformed self-exemplars},
  author={Huang, Jia-Bin and Singh, Abhishek and Ahuja, Narendra},
  booktitle={Proceedings of the IEEE conference on computer vision and pattern recognition},
  pages={5197--5206},
  year={2015}
}

@article{yang2019joint,
  title={Joint rain detection and removal from a single image with contextualized deep networks},
  author={Yang, Wenhan and Tan, Robby T and Feng, Jiashi and Guo, Zongming and Yan, Shuicheng and Liu, Jiaying},
  journal={IEEE transactions on pattern analysis and machine intelligence},
  volume={42},
  number={6},
  pages={1377--1393},
  year={2019},
  publisher={IEEE}
}

@inproceedings{luo2015removing,
  title={Removing rain from a single image via discriminative sparse coding},
  author={Luo, Yu and Xu, Yong and Ji, Hui},
  booktitle={Proceedings of the IEEE international conference on computer vision},
  pages={3397--3405},
  year={2015}
}

@inproceedings{li2016rain,
  title={Rain streak removal using layer priors},
  author={Li, Yu and Tan, Robby T and Guo, Xiaojie and Lu, Jiangbo and Brown, Michael S},
  booktitle={Proceedings of the IEEE conference on computer vision and pattern recognition},
  pages={2736--2744},
  year={2016}
}

@inproceedings{gu2017joint,
  title={Joint convolutional analysis and synthesis sparse representation for single image layer separation},
  author={Gu, Shuhang and Meng, Deyu and Zuo, Wangmeng and Zhang, Lei},
  booktitle={Proceedings of the IEEE international conference on computer vision},
  pages={1708--1716},
  year={2017}
}

@article{fu2017clearing,
  title={Clearing the skies: A deep network architecture for single-image rain removal},
  author={Fu, Xueyang and Huang, Jiabin and Ding, Xinghao and Liao, Yinghao and Paisley, John},
  journal={IEEE Transactions on Image Processing},
  volume={26},
  number={6},
  pages={2944--2956},
  year={2017},
  publisher={IEEE}
}

@inproceedings{fu2017removing,
  title={Removing rain from single images via a deep detail network},
  author={Fu, Xueyang and Huang, Jiabin and Zeng, Delu and Huang, Yue and Ding, Xinghao and Paisley, John},
  booktitle={Proceedings of the IEEE conference on computer vision and pattern recognition},
  pages={3855--3863},
  year={2017}
}

@inproceedings{li2018recurrent,
  title={Recurrent squeeze-and-excitation context aggregation net for single image deraining},
  author={Li, Xia and Wu, Jianlong and Lin, Zhouchen and Liu, Hong and Zha, Hongbin},
  booktitle={Proceedings of the European conference on computer vision (ECCV)},
  pages={254--269},
  year={2018}
}

@inproceedings{ren2019progressive,
  title={Progressive image deraining networks: A better and simpler baseline},
  author={Ren, Dongwei and Zuo, Wangmeng and Hu, Qinghua and Zhu, Pengfei and Meng, Deyu},
  booktitle={Proceedings of the IEEE/CVF conference on computer vision and pattern recognition},
  pages={3937--3946},
  year={2019}
}

@inproceedings{wang2019spatial,
  title={Spatial attentive single-image deraining with a high quality real rain dataset},
  author={Wang, Tianyu and Yang, Xin and Xu, Ke and Chen, Shaozhe and Zhang, Qiang and Lau, Rynson WH},
  booktitle={Proceedings of the IEEE/CVF conference on computer vision and pattern recognition},
  pages={12270--12279},
  year={2019}
}

@inproceedings{wei2019semi,
  title={Semi-supervised transfer learning for image rain removal},
  author={Wei, Wei and Meng, Deyu and Zhao, Qian and Xu, Zongben and Wu, Ying},
  booktitle={Proceedings of the IEEE/CVF conference on computer vision and pattern recognition},
  pages={3877--3886},
  year={2019}
}

@article{xiao2022image,
  title={Image de-raining transformer},
  author={Xiao, Jie and Fu, Xueyang and Liu, Aiping and Wu, Feng and Zha, Zheng-Jun},
  journal={IEEE Transactions on Pattern Analysis and Machine Intelligence},
  volume={45},
  number={11},
  pages={12978--12995},
  year={2022},
  publisher={IEEE}
}

@inproceedings{hertel2015deep,
  title={Deep convolutional neural networks as generic feature extractors},
  author={Hertel, Lars and Barth, Erhardt and K{\"a}ster, Thomas and Martinetz, Thomas},
  booktitle={2015 International Joint Conference on Neural Networks (IJCNN)},
  pages={1--4},
  year={2015},
  organization={IEEE}
}

@inproceedings{albahar2019guided,
  title={Guided image-to-image translation with bi-directional feature transformation},
  author={AlBahar, Badour and Huang, Jia-Bin},
  booktitle={Proceedings of the IEEE/CVF international conference on computer vision},
  pages={9016--9025},
  year={2019}
}

@inproceedings{shaham2021spatially,
  title={Spatially-adaptive pixelwise networks for fast image translation},
  author={Shaham, Tamar Rott and Gharbi, Micha{\"e}l and Zhang, Richard and Shechtman, Eli and Michaeli, Tomer},
  booktitle={Proceedings of the IEEE/CVF conference on computer vision and pattern recognition},
  pages={14882--14891},
  year={2021}
}

@article{kang2024deep,
  title={Deep convolutional dictionary learning network for sparse view CT reconstruction with a group sparse prior},
  author={Kang, Yanqin and Liu, Jin and Wu, Fan and Wang, Kun and Qiang, Jun and Hu, Dianlin and Zhang, Yikun},
  journal={Computer Methods and Programs in Biomedicine},
  volume={244},
  pages={108010},
  year={2024},
  publisher={Elsevier}
}

@article{xie2020mhf,
  title={MHF-Net: An interpretable deep network for multispectral and hyperspectral image fusion},
  author={Xie, Qi and Zhou, Minghao and Zhao, Qian and Xu, Zongben and Meng, Deyu},
  journal={IEEE Transactions on Pattern Analysis and Machine Intelligence},
  volume={44},
  number={3},
  pages={1457--1473},
  year={2020},
  publisher={IEEE}
}

@article{liu2024infrared,
  title={Infrared Small Target Detection via Joint Low Rankness and Local Smoothness Prior},
  author={Liu, Pei and Peng, Jiangjun and Wang, Hailin and Hong, Danfeng and Cao, Xiangyong},
  journal={IEEE Transactions on Geoscience and Remote Sensing},
  year={2024},
  publisher={IEEE}
}

@article{tan2025ds,
  title={DS-Net: A model driven network framework for lesion segmentation on fundus image},
  author={Tan, Feiyu and Wang, Yuhan and Xie, Qi and Fu, Jiahong and Wang, Renzhen and Meng, Deyu},
  journal={Knowledge-Based Systems},
  volume={315},
  pages={113242},
  year={2025},
  publisher={Elsevier}
}

@inproceedings{elesedy2022group,
  title={Group symmetry in pac learning},
  author={Elesedy, Bryn},
  booktitle={ICLR 2022 workshop on geometrical and topological representation learning},
  year={2022}
}

@inproceedings{liu2025rotation,
  title={Rotation-Equivariant Self-Supervised Method in Image Denoising},
  author={Liu, Hanze and Fu, Jiahong and Xie, Qi and Meng, Deyu},
  booktitle={Proceedings of the Computer Vision and Pattern Recognition Conference},
  pages={12720--12730},
  year={2025}
}

@article{xie2025rotation,
  title={Rotation Equivariant Arbitrary-scale Image Super-Resolution},
  author={Xie, Qi and Fu, Jiahong and Xu, Zongben and Meng, Deyu},
  journal={arXiv preprint arXiv:2508.05160},
  year={2025}
}

\vspace{-5mm}
\begin{IEEEbiography}[{\includegraphics[width=1in,height=1.25in,clip,keepaspectratio]{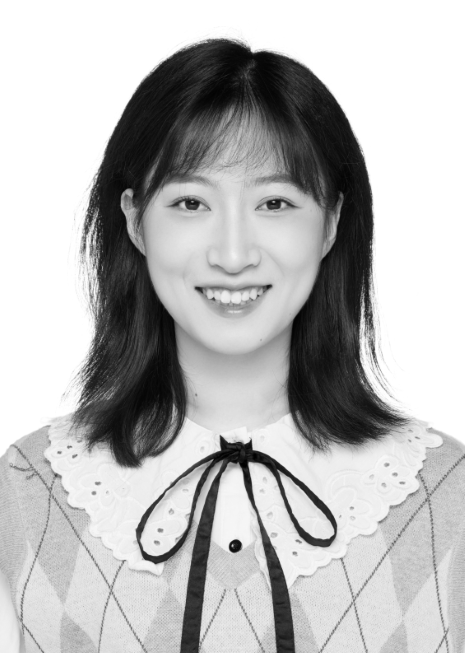}}]{Feiyu Tan} received the M.Sc. degree from Xi'an Jiaotong University, Xi'an, China, in 2024. She is currently pursuing the Ph.D. degree in Xi'an Jiaotong University.
Her current research interests include model-based deep learning and rotation equivariant deep learning.
\end{IEEEbiography}

\begin{IEEEbiography}[{\includegraphics[width=1in,height=1.25in,clip,keepaspectratio]{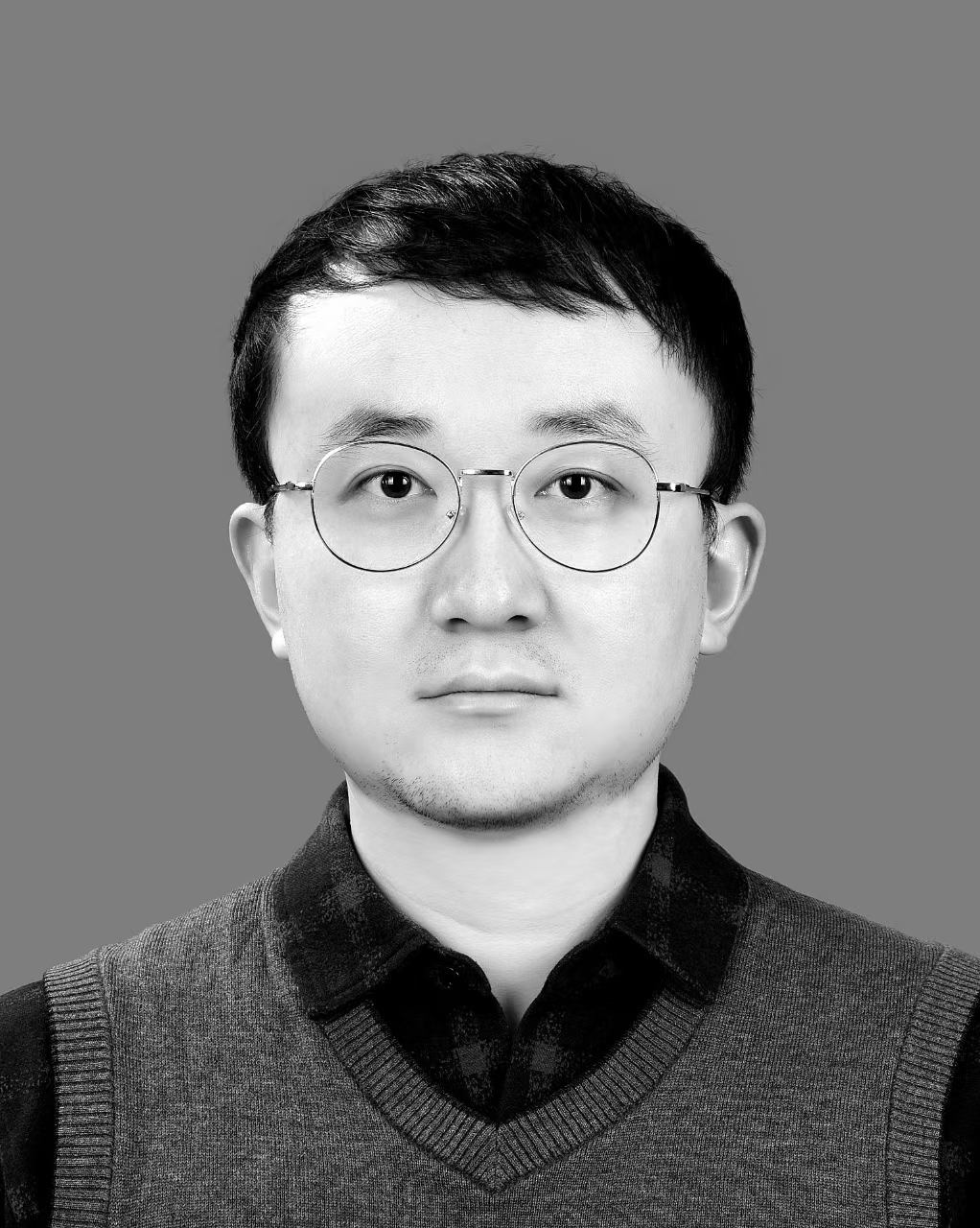}}]{Heran Yang} received the B.Sc. and Ph.D degree from Xi'an Jiaotong University, Xi'an, China, in 2013 and 2021 respectively. He was a Visiting Scholar with Johns Hopkins University, Baltimore, MD, USA, from 2017 to 2018. He is currently an associate professor with School of Mathematics and Statistics, Xi'an Jiaotong University. His current research interests include machine learning and medical image analysis.
\end{IEEEbiography}

\begin{IEEEbiography}[{\includegraphics[width=1in,height=1.25in,clip,keepaspectratio]{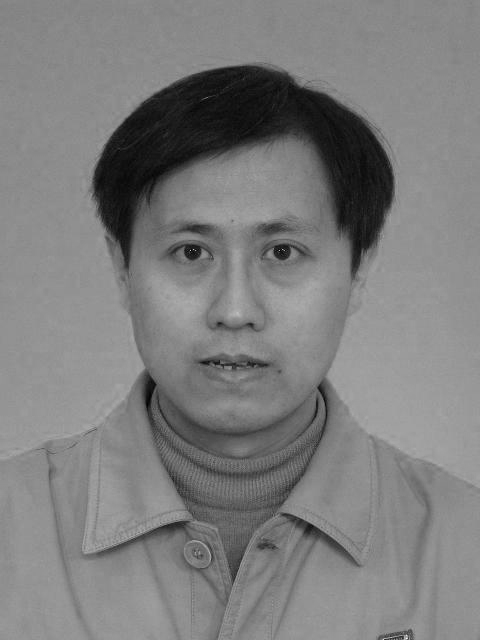}}]{Qihong Duan} received the B.S. and M.S. degrees in computational mathematics, and the Ph.D. degree in probability and statistics from Xi'an Jiaotong University, in 1993, 1996, and 2001, respectively. He is currently working as an Associate Professor at Xi'an Jiaotong University. His research interests include stochastic processes and probability model.
\end{IEEEbiography}

\begin{IEEEbiography}[{\includegraphics[width=1in,height=1.25in,clip,keepaspectratio]{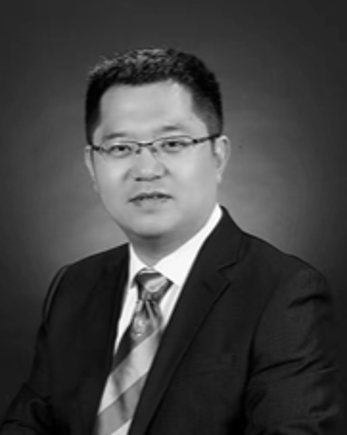}}]{Kai Ye} received the B.S. and M.S. degrees from Wuhan University, Wuhan, China, and the Ph.D. degree from Leiden University, Leiden, The Netherlands, in 2008. After one year postdoctoral training with European Bioinformatics Institute, Cambridge, U.K., he joined Leiden University Medical Center as an Assistant Professor in 2009. In 2012, he moved to Washington University in St. Louis, St. Louis, MO, USA. He is a Full Professor with the School of Electronics and Information Engineering, Xi’an Jiaotong University, Xi’an, China.  His research interests include sequential pattern mining, computational methodology on biomedical big data, and applications on precision medicine.
\end{IEEEbiography}

\begin{IEEEbiography}[{\includegraphics[width=1in,height=1.25in,clip,keepaspectratio]{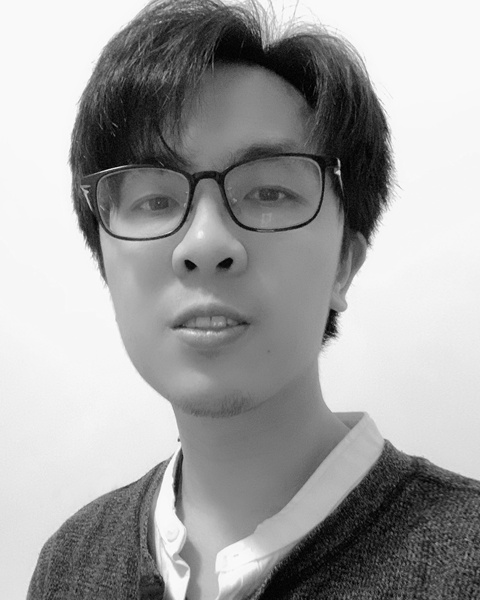}}]{Qi Xie} received the B.Sc. and Ph.D degrees from Xi'an Jiaotong University, Xi'an, China, in 2013 and 2020 respectively.  Form 2018 to 2019, he was a Visiting Scholar in Princeton University, Princeton, NJ, USA. 
He is currently an associate professor in School of Mathematics and Statistics, Xi'an Jiaotong University.
His current research interests include model-based deep learning, filter parametrization-based deep learning, rotation equivariant deep learning.
\end{IEEEbiography}

\begin{IEEEbiography}[{\includegraphics[width=1in,height=1.25in,clip,keepaspectratio]{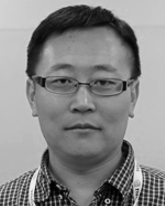}}]{Deyu Meng} received the B.Sc., M.Sc., and Ph.D. degrees from Xi'an Jiaotong University, Xi'an, China, in 2001, 2004, and 2008, respectively. He is currently a professor in School of Mathematics and Statistics, Xi'an Jiaotong University, and adjunct professor in Faculty of Information Technology, The Macau University of Science and Technology. From 2012 to 2014, he took his two-year sabbatical leave in Carnegie Mellon University. His current research interests include machine learning and meta-learning.
\end{IEEEbiography}

\end{document}